\documentclass{article}

\usepackage{titling}
\usepackage{subfiles}

\PassOptionsToPackage{numbers, compress}{natbib}

\usepackage{xr}
\usepackage[utf8]{inputenc} 
\usepackage[T1]{fontenc}    
\usepackage{hyperref}       
\usepackage{url}            
\usepackage{booktabs}       
\usepackage{amsfonts}       
\usepackage{nicefrac}       
\usepackage{microtype}      
\usepackage{graphicx}
\usepackage{amsmath}
\usepackage{xcolor}
\usepackage{mathtools}
\usepackage{caption}
\usepackage{subcaption}
\usepackage{wrapfig}
\usepackage{verbatim}
\usepackage[verbose]{placeins}
\usepackage{authblk}
\usepackage{cite}
\usepackage[verbose=true,letterpaper]{geometry}

\DeclareMathOperator*{\argmax}{arg\,max}

\DeclareMathOperator*{\R}{\mathbb{R}}

\newcommand{\relu}{\text{relu}}
\newcommand{\softplus}{\text{softplus}}

\newcommand{\dg}{\dot{\gamma}}
\newtheorem{theorem}{Theorem}

\newcounter{oldtocdepth}

\newcommand{\hidefromtoc}{%
  \setcounter{oldtocdepth}{\value{tocdepth}}%
  \addtocontents{toc}{\protect\setcounter{tocdepth}{-10}}%
}

\newcommand{\unhidefromtoc}{%
  \addtocontents{toc}{\protect\setcounter{tocdepth}{\value{oldtocdepth}}}%
}

\newgeometry{
    textheight=9in,
    textwidth=5.5in,
    top=1in,
    headheight=12pt,
    headsep=25pt,
    footskip=30pt
}
\begin{document}

\hidefromtoc

\maketitle
\begin{abstract}
Explanation methods aim to make neural networks more trustworthy and interpretable. In this paper, we demonstrate a property of explanation methods which is disconcerting for both of these purposes. Namely, we show that explanations can be manipulated \emph{arbitrarily} by applying visually hardly perceptible perturbations to the input that keep the network's output approximately constant. We establish theoretically that this phenomenon can be related to certain geometrical properties of neural networks. This allows us to derive an upper bound on the susceptibility of explanations to manipulations. Based on this result, we propose effective mechanisms to enhance the robustness of explanations.
\end{abstract}

\begin{figure}[h!]
  \centering
  \includegraphics[width=0.4\linewidth]{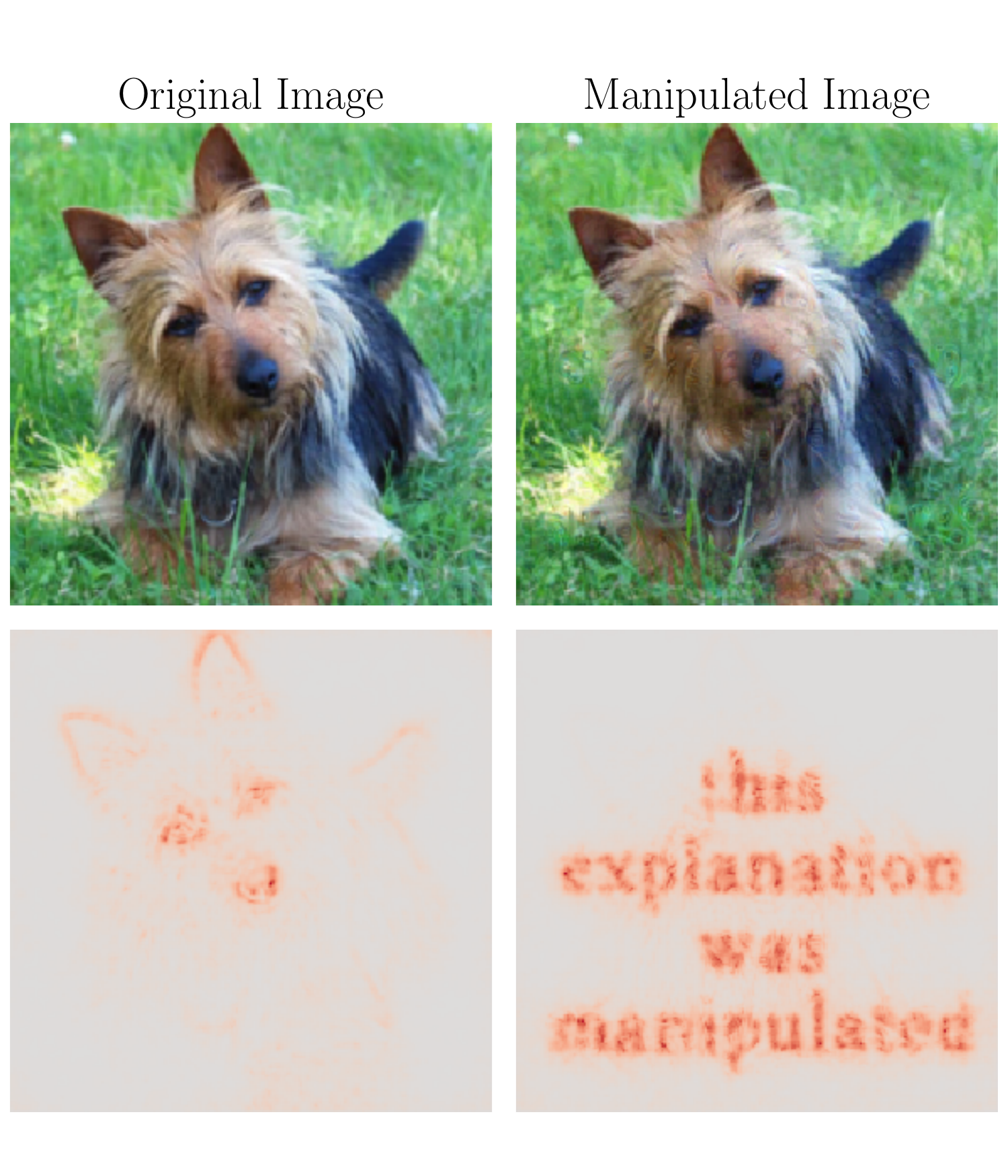}
  \caption{Original image with corresponding explanation map on the left. Manipulated image with its explanation on the right. The chosen target explanation was an image with a text stating "this explanation was manipulated". \label{fig:aufmacher}}
\end{figure}

\section{Introduction}
Explanation methods have attracted significant attention over the last years due to their promise to open the black box of deep neural networks. Interpretability is crucial for scientific understanding and safety critical applications.

Explanations can be provided in terms of explanation maps\cite{grad2, grad1, expl_2, gbp, lrp, gradcam, expl_8, expl_3, expl_4, expl_5, expl_6, expl_9, integratedgrad, gradtimesinput, fong2017interpretable, dtd, patternattr, kim2017interpretability, lapuschkin2019unmasking}  that visualize the relevance attributed to each input feature for the overall classification result. In this work, we establish that these explanation maps can be changed to an \emph{arbitrary target map}. This is done by applying a visually hardly perceptible perturbation to the input. We refer to Figure~\ref{fig:aufmacher} for an example. This perturbation does not change the output of the neural network, i.e. in addition to the classification result also the vector of all class probabilities is (approximately) the same. 

This finding is clearly problematic if a user, say a medical doctor, is expecting a robustly interpretable explanation map to rely on in the clinical decision making process. 

Motivated by this unexpected observation, we provide a theoretical analysis that establishes a relation of this phenomenon to the geometry of the neural network's output manifold. This novel understanding allows us to derive a bound on the degree of possible manipulation of the explanation map. This bound is proportional to two differential geometric quantities: the principle curvatures and the geodesic distance between the original input and its manipulated counterpart. Given this theoretical insight, we propose efficient ways to limit possible manipulations and thus enhance resilience of explanation methods.

In summary, this work provides the following key contributions:
\begin{itemize}
    \item We propose an algorithm which allows to manipulate an image with a hardly perceptible perturbation such that the explanation matches an arbitrary target map. We demonstrate its effectiveness for six different explanation methods and on four network architectures as well as two datasets.
  \item We provide a theoretical understanding of this phenomenon for gradient-based methods in terms of differential geometry. We derive a bound on the principle curvatures of the hypersurface of equal network output. This implies a constraint on the maximal change of the explanation map due to small perturbations. 
  \item Using these insights, we propose methods to undo the manipulations and increase the robustness of explanation maps by smoothing the explanation method. We demonstrate experimentally that smoothing leads to increased robustness not only for gradient but also for propagation-based methods. 
\end{itemize}

\subsection{Related work}
In \cite{fragile}, it was demonstrated that explanation maps can be sensitive to small perturbations in the image. Their results may be thought of as untargeted manipulations, i.e. perturbations to the image which lead to an unstructured change in the explanation map. Our work focuses on targeted manipulations instead, i.e. to reproduce a given target map. Another approach \cite{cat} adds a constant shift to the input image, which is then eliminated by changing the bias of the first layer. For some methods, this leads to a change in the explanation map. Contrary to our approach, this requires to change the network's biases.
In \cite{sanity}, explanation maps are changed by randomization of (some of) the network weights. This is different from our method as it does not aim to change the explanation in a targeted manner and modifies the weights of the network.

\section{Manipulating explanations}
We consider a neural network $g: \mathbb{R}^d \to \mathbb{R}^K$ with  $\text{relu}$ non-linearities which classifies an image $x \in \mathbb{R}^d$ in $K$ categories with the predicted class given by $k=\argmax_{i} g(x)_i$. The explanation map is denoted by $h: \mathbb{R}^d \to \mathbb{R}^d$ and associates an image with a vector of the same dimension whose components encode the relevance score of each pixel for the neural network's prediction. 
For a given explanation method and specified target $h^t\in \mathbb{R}^d$, a manipulated image $x_{adv} = x + \delta x$ has the following properties:
\begin{enumerate}
    \item The output of the network stays approximately constant, i.e.  $g(x_{adv}) \approx g(x)$.
    \item The explanation is close to the target map, i.e. $h(x_{adv}) \approx h^t$.
    \item The norm of the perturbation $\delta x$ added to the input image is small, i.e. $\left\|\delta x\right\|=\left\|x_{adv}- x\right\| \ll 1$ and therefore not perceptible.
\end{enumerate}
Throughout this paper, we will use the following explanation methods:
\begin{itemize}
    \item \textbf{Gradient}: The map $h(x)=\frac{\partial g}{\partial x}(x)$ is used and quantifies how infinitesimal perturbations in each pixel change the prediction $g(x)$ \cite{grad1,grad2}.
    \item \textbf{Gradient $\times$ Input}: This method uses the map $h(x)=x \odot \frac{\partial g}{\partial x}(x)$  \cite{gradtimesinput}. For linear models, this measure gives the exact contribution of each pixel to the prediction.
    \item \textbf{Integrated Gradients}: This method defines $h(x)= (x-\bar{x}) \odot \int_0^1 \frac{\partial g(\bar{x}+t(x-\bar{x}))}{\partial x}\text{d}t$ where $\bar{x}$ is a suitable baseline. See the original reference \cite{integratedgrad} for more details.
    \item \textbf{Guided Backpropagation (GBP)}: This method is a variation of the gradient explanation for which negative components of the gradient are set to zero while backpropagating through the non-linearities \cite{gbp}.
    \item \textbf{Layer-wise Relevance Propagation (LRP)}: This method \cite{lrp, dtd} propagates relevance backwards through the network. For the output layer, relevance is defined by\footnote{Here we use the Kronecker symbol 
    $\delta_{i,k}= \begin{cases}
    1, & \text{for } i = k  \\
    0, & \text{for } i \neq k
    \end{cases}$.} 
    \begin{align}
        R^L_i = \delta_{i,k} \,, \label{eq:lrp_final_layer}
    \end{align}
    which is then propagated backwards through all layers but the first using the $z^+$ rule
    \begin{align}
        R^l_i = \sum_{j} \frac{x_i^l (W^l)^+_{ji}}{\sum_i x_i^l (W^l)^+_{ji}} R^{l+1}_j \,, \label{eq:lrp_interm_layer}
    \end{align}
    where $(W^l)^+$ denotes the positive weights of the $l$-th layer and $x^l$ is the activation vector of the $l$-th layer.
    For the first layer, we use the $z^\mathcal{B}$ rule to account for the bounded input domain 
    \begin{align}
        R^0_i = \sum_{j} \frac{x_j^0 W^{0}_{ji}-l_j (W^{0})^+_{ji}-h_j (W^{0})^{-}_{ji}}{\sum_i ( x_j^0 W^{0}_{ji}-l_j (W^{0})^+_{ji}-h_j (W^{0})^{-}_{ji})} R^{1}_j \,, \label{eq:lrp_first_layer}
    \end{align}
    where $l_i$ and $h_i$ are the lower and upper bounds of the input domain respectively.
    \item \textbf{Pattern Attribution (PA)}: This method is equivalent to standard backpropagation upon element-wise multiplication of the weights $W^l$ with learned patterns $A^l$. We refer to the original publication for more details \cite{patternattr}.
\end{itemize}
These methods cover two classes of attribution methods, namely \emph{gradient-based} and \emph{propagation-based} explanations, and are frequently used in practice \cite{innvestigate, deeplift}. 

\subsection{Manipulation Method}
Let $h^t \in \mathbb{R}^d$ be a given target explanation map and $x \in  \mathbb{R}^d$ an input image. As explained previously, we want to construct a manipulated image $x_{adv}=x+\delta x$ such that it has an explanation very similar to the target $h^t$ but the output of the network stays approximately constant, i.e. $g(x_{adv}) \approx g(x)$. We obtain such manipulations by optimizing the loss function
\begin{align}
    \mathcal{L} = \left\| h(x_{adv})-h^t\right\|^2 + \gamma \, \left\| g(x_{adv})-g(x)\right\|^2 \,, \label{eq:loss}
\end{align}
with respect to $x_{adv}$ using gradient descent. We clamp $x_{adv}$ after each iteration so that it is a valid image. The first term in the loss function \eqref{eq:loss} ensures that the manipulated explanation map is close to the target while the second term encourages the network to have the same output. The relative weighting of these two summands is controlled by the hyperparameter $\gamma\in\mathbb{R}_+$.

\begin{figure*}[t!]
  \centering
  \includegraphics[width=1.0\linewidth]{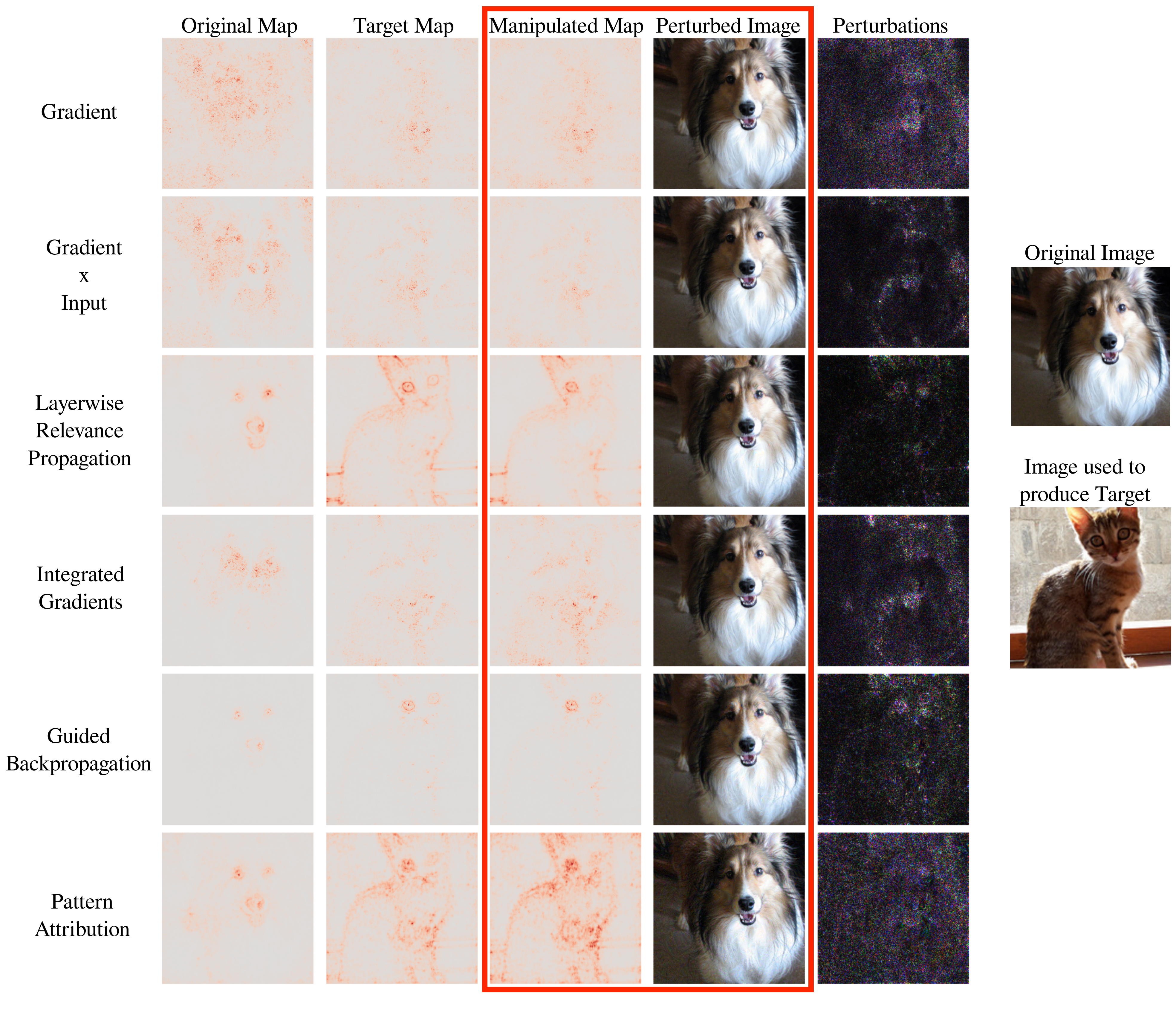}
  \caption{The explanation map of the cat is used as the target and the image of the dog is perturbed. 
The red box contains the manipulated images and the corresponding explanations. The first column corresponds to the original explanations of the unperturbed dog image. The target map, shown in the second column, is generated with the cat image.  The last column visualizes the perturbations. \label{fig:masterplot}}
\end{figure*}

\begin{figure}[h!]
  \centering
  \includegraphics[width=1.0\linewidth]{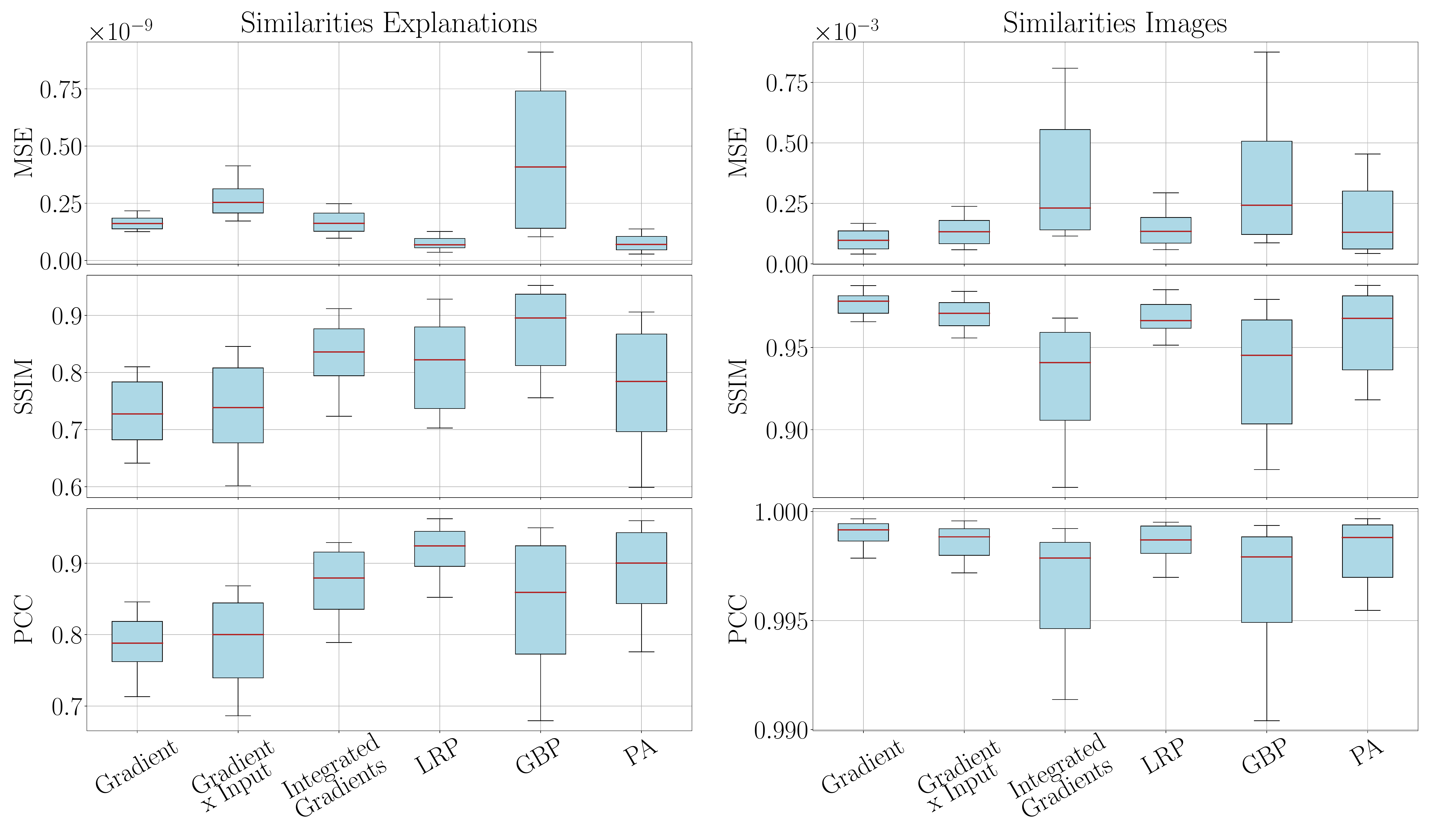}
  \caption{Left: Similarity measures between target $h^t$ and manipulated explanation map $h(x_{adv})$. Right: Similarity measures between original $x$ and perturbed image $x_{adv}$. For SSIM and PCC large values indicate high similarity while for MSE small values correspond to similar images.\label{fig:quantitative}}
\end{figure}

The gradient with respect to the input $\nabla h(x)$ of the explanation often depends on the vanishing second derivative of the $\text{relu}$ non-linearities. This causes problems during optimization of the loss \eqref{eq:loss}. As an example, the gradient method leads to
\begin{align*}
\partial_{x_{adv}} \left\| h(x_{adv})-h^t\right\|^2   \propto \frac{\partial h}{\partial x_{adv}} = \frac{\partial^2 g}{\partial x_{adv}^2} \propto \relu'' = 0 \,. 
\end{align*}
We therefore replace the $\relu$ by softplus non-linearities
\begin{align}
    \softplus_\beta(x) = \frac{1}{\beta} \log(1+e^{\beta x}) \,. \label{eq:softplus}
\end{align}
For large $\beta$ values, the softplus approximates the $\relu$ closely but has a well-defined second derivative. After optimization is complete, we test the manipulated image with the original $\relu$ network.

\textbf{Similarity metrics:} In our analysis, we assess the similarity between both images and explanation maps. To this end, we use three metrics following \cite{sanity}: the structural similarity index (SSIM), the Pearson correlation coefficient (PCC) and the mean squared error (MSE). SSIM and PCC are relative similarity measures with values in $[0,1]$, where larger values indicate high similarity. The MSE is an absolute error measure for which values close to zero indicate high similarity. We normalize the sum of the explanation maps to be one and the images to have values between 0 and 1. 

\subsection{Experiments}\label{sec:experiments}
To evaluate our approach, we apply our algorithm to 100 randomly selected images for each explanation method. We use a pre-trained VGG-16 network \cite{vgg} and the ImageNet dataset \cite{imagenet}. For each run, we randomly select two images from the test set. One of the two images is used to generate a target explanation map $h^t$. The other image is perturbed by our algorithm with the goal of replicating the target $h^t$ using a few thousand iterations of gradient descent. We sum over the absolute values of the channels of the explanation map to get the relevance per pixel. Further details about the experiments are summarized in Supplement~\ref{app:details_experiments}.

\textbf{Qualitative analysis:} Our method is illustrated in Figure~\ref{fig:masterplot} in which a dog image is manipulated in order to have an explanation of a cat. For all explanation methods, the target is closely emulated and the perturbation of the dog image is small. More examples can be found in the supplement.

\textbf{Quantitative analysis:} Figure~\ref{fig:quantitative} shows similarity measures between the target $h^t$ and the manipulated explanation map $h(x_{adv})$ as well as between the original image $x$ and perturbed image $x_{adv}$.\footnote{Throughout this paper, boxes denote 25\textsuperscript{th} and 75\textsuperscript{th} percentiles, whiskers denote 10\textsuperscript{th} and 90\textsuperscript{th} percentiles, solid lines show the medians and outliers are depicted by circles.} All considered metrics show that the perturbed images have an explanation closely resembling the targets. At the same time, the perturbed images are very similar to the corresponding original images.  We also verified by visual inspection that the results look very similar. We have uploaded the results of all runs so that interested readers can assess their similarity themselves\footnote{\url{https://drive.google.com/drive/folders/1TZeWngoevHRuIw6gb5CZDIRrc7EWf5yb?usp=sharing}} and will provide code to reproduce them. In addition, the output of the neural network is approximately unchanged by the perturbations, i.e. the classification of all examples is unchanged and the median of $\left\| g(x_{adv})-g(x)\right\|$ is of the order of magnitude $10^{-3}$ for all methods. See Supplement~\ref{app:diff_output} for further details.

\textbf{Other architectures and datasets:}
We checked that comparable results are obtained for ResNet-18 \cite{resnet}, AlexNet \cite{alex} and Densenet-121 \cite{densenet}. Moreover, we also successfully tested our algorithm on the CIFAR-10 dataset \cite{cifar}. We refer to the Supplement~\ref{app:other_architectures_datasets} for further details.

\section{Theoretical considerations}
In this section, we analyze the vulnerability of explanations theoretically. We argue that this phenomenon can be related to the large curvature of the output manifold of the neural network. We focus on the gradient method starting with an intuitive discussion before developing mathematically precise statements.

We have demonstrated that one can drastically change the explanation map while keeping the output of the neural network constant 
\begin{align}
    g(x+\delta x) = g(x) = c
\end{align}
using only a small perturbation in the input $\delta x$. The perturbed image $x_{adv}=x+\delta x$ therefore lies on the hypersurface of constant network output $S = \{ p \in \mathbb{R}^d | g(p)=c \}$.\footnote{It is sufficient to consider the hypersurface $S$ in a neighbourhood of the unperturbed input $x$.} We can exclusively consider the winning class output, i.e. $g(x):=g(x)_k$ with $k=\argmax_{i} g(x)_i$ because the gradient method only depends on this component of the output. Therefore, the hyperplane $S$ is of co-dimension one. The gradient $\nabla g$ for every $p \in S$ is normal to this hypersurface. The fact that the normal vector $\nabla g$ can be drastically changed by slightly perturbing the input along the hypersurface $S$ suggests that the curvature of $S$ is large.

While the latter statement may seem intuitive, it requires non-trivial concepts of differential geometry to make it precise, in particular the notion of the second fundamental form. We will briefly summarize these concepts in the following (see e.g. \cite{tu2017differential} for a standard textbook). To this end, it is advantageous to consider a normalized version of the gradient method
\begin{align}
   n(x) = \frac{\nabla g(x)}{\left\|\nabla g(x)\right\|} \label{eq:normal} \,.
\end{align}
This normalization is merely conventional as it does not change the relative importance of any pixel with respect to the others. For any point $p \in S$, we define the tangent space $T_p S$ as the vector space spanned by the tangent vectors $\dg(0)=\tfrac{d}{dt} \gamma(t) |_{t=0}$ of all possible curves $\gamma: \mathbb{R} \to S$ with $\gamma(0)=p$. For $u,v \in T_p S$, we denote their inner product by $\langle u,v \rangle$. For any $u \in T_p S$, the \emph{directional derivative} is uniquely defined for any choice of $\gamma$ by
\begin{align}
    D_u f(p) = \left. \frac{d}{dt} f(\gamma(t)) \right|_{t=0} && \text{with} && \gamma(0)=p \; \; \text{and} \; \;  \dg(0)=u.
\end{align}
We then define the \emph{Weingarten map} as\footnote{The fact that $D_u n(p) \in T_p S$ follows by taking the directional derivative with respect to $u$ on both sides of $\langle n, n \rangle = 1$ .}
\begin{align}
L:  \, \begin{cases}
T_p S &\to  \;T_p S \nonumber \\
 u &\mapsto \;  -D_u n(p) \,,
 \end{cases}
\end{align}
where the unit normal $n(p)$ can be written as \eqref{eq:normal}.
This map quantifies how much the unit normal changes as we infinitesimally move away from $p$ in the direction $u$. The \emph{second fundamental form} is then given by 
\begin{align}
\mathcal{L}:  \, \begin{cases}
T_p S \times T_p S  &\to  \;\mathbb{R} \nonumber \\
 u, v &\mapsto \;  -\langle v, L(u) \rangle =  -\langle v, D_u n(p) \rangle \,.
 \end{cases}
\end{align}
It can be shown that the second fundamental form is bilinear and symmetric $\mathcal{L}(u,v) = \mathcal{L}(v,u)$. It is therefore diagonalizable with real eigenvalues $\lambda_1, \dots \lambda_{d-1}$ which are called \emph{principle curvatures}. 

We have therefore established the remarkable fact that the sensitivity of the gradient map \eqref{eq:normal} is described by the principle curvatures, a key concept of differential geometry.

In particular, this allows us to derive an upper bound on the maximal change of the gradient map $h(x)=n(x)$  as we move slightly on $S$. To this end, we define the \emph{geodesic distance} $d_g(p,q)$ of two points $p,q \in S$ as the length of the shortest curve on $S$ connecting $p$ and $q$. In the supplement, we show that:

\begin{theorem}
Let $g: \mathbb{R}^d \to \mathbb{R}$ be a network with $\softplus_\beta$ non-linearities and $\mathcal{U}_\epsilon(p) = \{ x \in \mathbb{R}^d ; \left\| x - p \right\| < \epsilon \}$ an environment of a point $p\in S$ such that $\mathcal{U}_\epsilon(p) \cap S$ is fully connected. Let $g$ have bounded derivatives $\left\|\nabla g(x)\right\|\ge c$ for all $x\in \mathcal{U}_\epsilon(p) \cap S$. It then follows for all $p_0 \in \mathcal{U}_\epsilon(p) \cap S$ that
\begin{align}
    \left\| h(p) - h(p_0) \right\| \le |\lambda_{max}| \; d_g(p, p_0) \le \beta \, C \, d_g(p, p_0),
\end{align}
where $\lambda_{max}$ is the principle curvature with the largest absolute value for any point in $\mathcal{U}_\epsilon(p) \cap S$ and the constant $C>0$ depends on the weights of the neural network.
\end{theorem}

This theorem can intuitively be motivated as follows: for $\relu$ non-linearities, the lines of equal network output are piece-wise linear and therefore have kinks, i.e. points of divergent curvature. These $\relu$ non-linearities are well approximated by $\softplus$ non-linearities~\eqref{eq:softplus} with large $\beta$. Reducing $\beta$ smoothes out the kinks and therefore leads to reduced maximal curvature, i.e. $|\lambda_{max}| \le \beta \, C$. For each point on the geodesic curve connecting $p$ and $p_0$, the normal can at worst be affected by the maximal curvature, i.e. the change in explanation is bounded by $|\lambda_{max}| \; d_g(p, p_0)$. 
\begin{figure}[h!]
\centering
\begin{subfigure}{.17\textwidth}
  \includegraphics[width=1\linewidth]{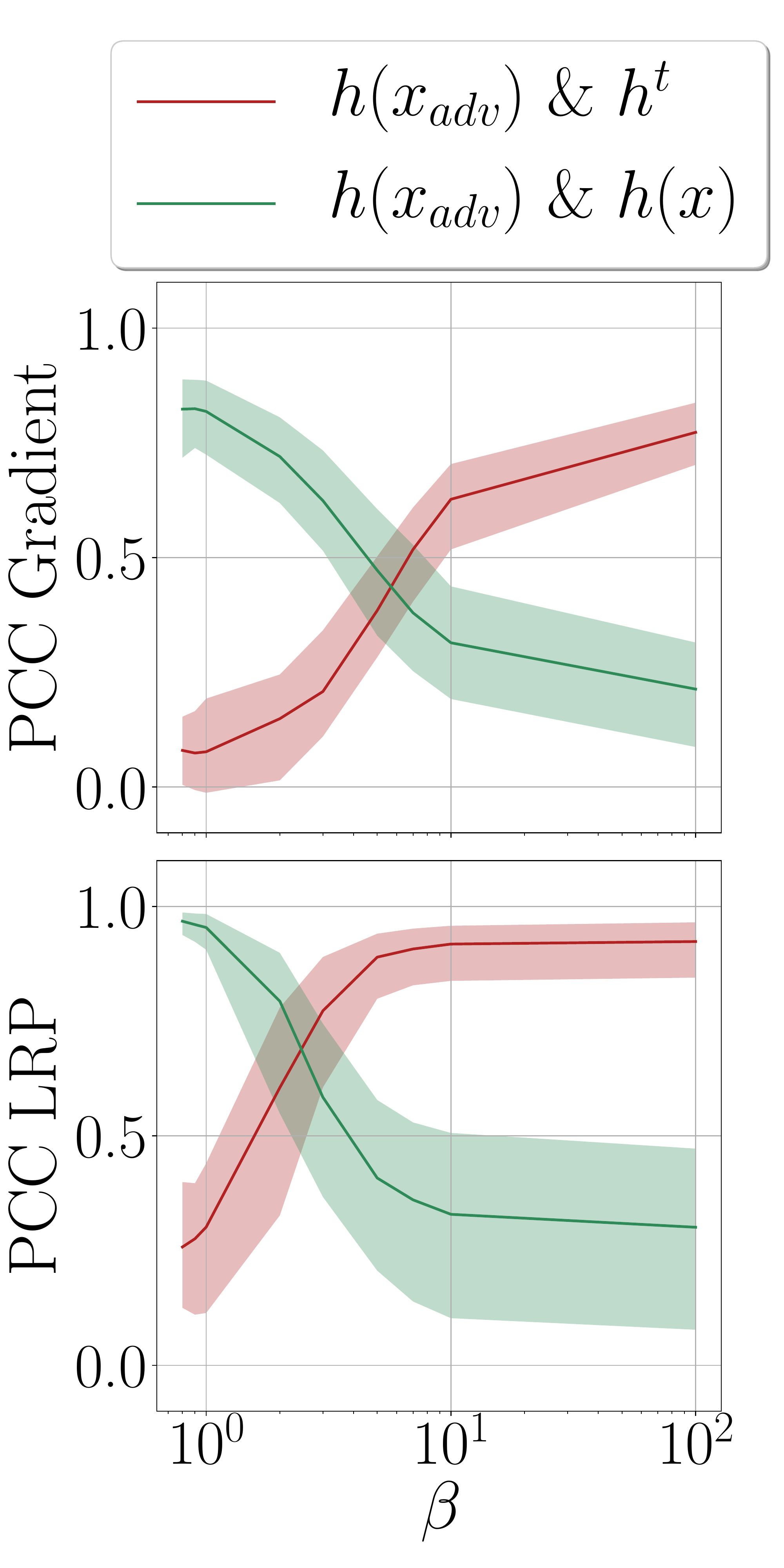}
\end{subfigure}%
\hfill
\begin{subfigure}{.4\textwidth}
  \includegraphics[width=1\linewidth]{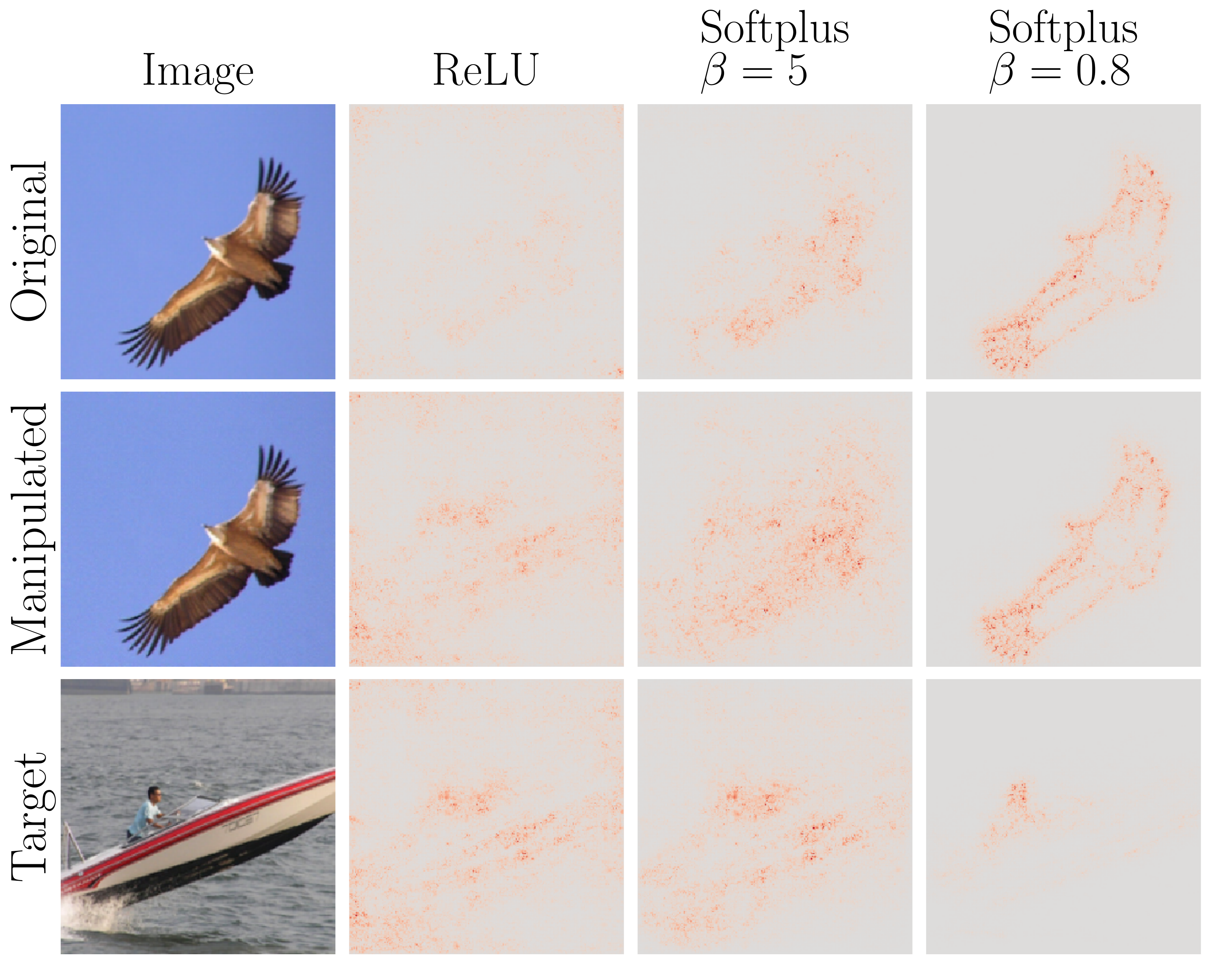}
 \end{subfigure}
 \begin{subfigure}{.4\textwidth}
  \includegraphics[width=1\linewidth]{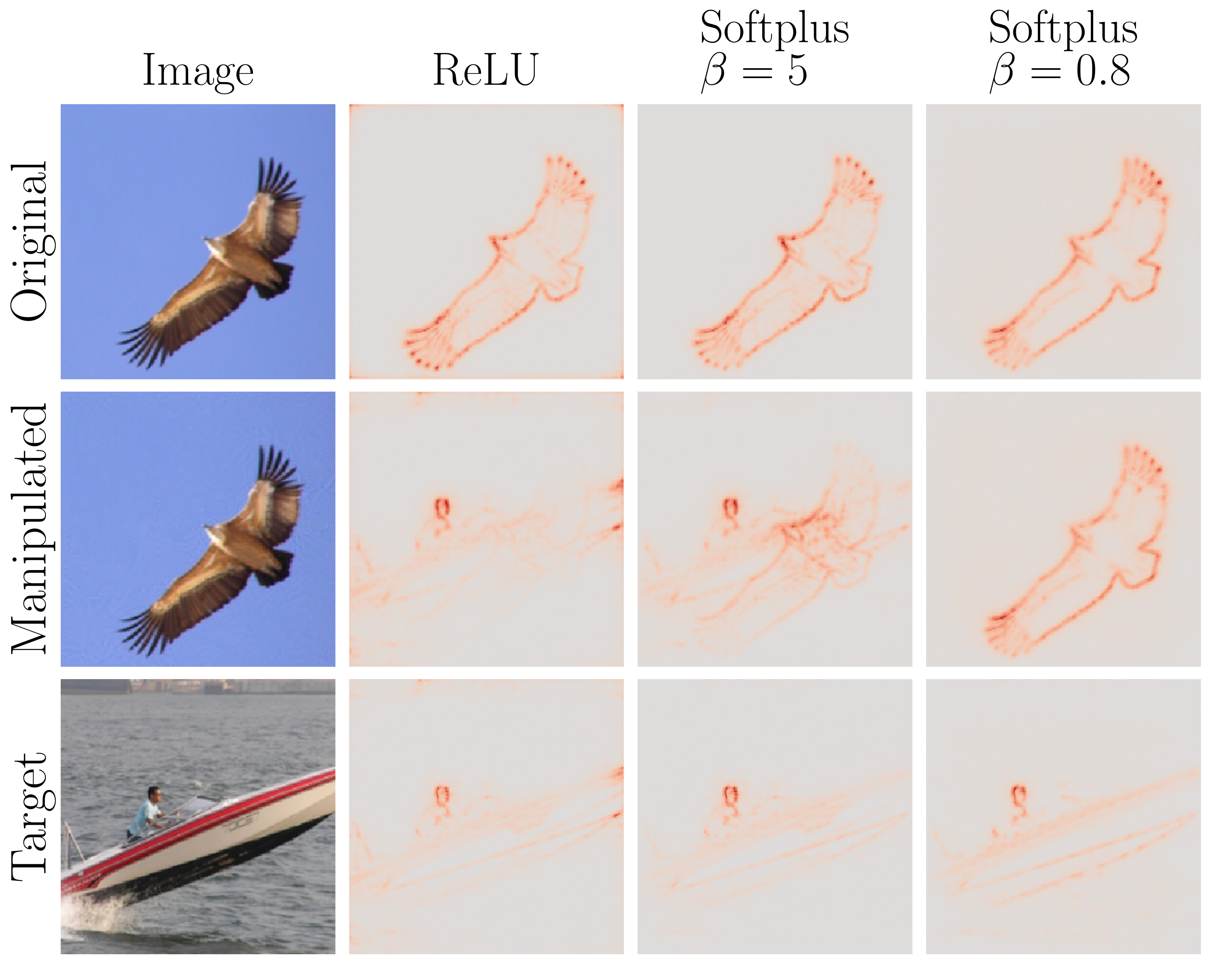}
 \end{subfigure}
 \caption{Left: $\beta$ dependence for the correlations of the manipulated explanation (here Gradient and LRP) with the target and original explanation. Lines denote the medians, $10^{th}$ and $90^{th}$ percentiles are shown in semitransparent colour. Center and Right: network input and the respective explanation maps as $\beta$ is decreased for Gradient (center) and LRP (right). }
 \label{fig:softplus}
\end{figure}

There are two important lessons to be learned from this theorem: the geodesic distance can be substantially greater than the Euclidean distance for curved manifolds. In this case, inputs which are very similar to each other, i.e. the Euclidean distance is small, can have explanations that are drastically different.
Secondly, the upper bound is proportional to the $\beta$ parameter of the softplus non-linearity. Therefore, smaller values of $\beta$ provably result in increased robustness with respect to manipulations.

\section{Robust explanations}

Using the fact that the upper bound of the last section is proportional to the $\beta$ parameter of the softplus non-linearities, we propose \emph{$\beta$-smoothing} of explanations. This method calculates an explanation using a network for which the $\relu$ non-linearities are replaced by $\softplus$ with a small $\beta$ parameter to smooth the principle curvatures. The precise value of $\beta$ is a hyperparameter of the method, but we find that a value around one works well in practice.

As shown in the supplement, a relation between SmoothGrad~\cite{expl_9} and $\beta$-smoothing can be proven for a one-layer neural network:
\begin{theorem}
For a one-layer neural network $g(x)=\relu(w^T x)$ and its $\beta$-smoothed counterpart  $g_{\beta}(x)=\softplus_\beta(w^T x)$, it holds that
\begin{align*}
   \mathbb{E}_{\epsilon \sim p_{\beta}} \left[ \nabla g(x-\epsilon) \right] = \nabla g_{\frac{\beta}{\left\| w \right\|}}(x) \,, 
\end{align*}
where $p_\beta(\epsilon) = \frac{\beta}{(e^{\beta \epsilon/2}+e^{-\beta \epsilon/2})^2}$.
\end{theorem}
\begin{wrapfigure}[9]{r}{0.3\textwidth}
\vspace{-17pt}
  \begin{center}
    \includegraphics[width=0.28\textwidth]{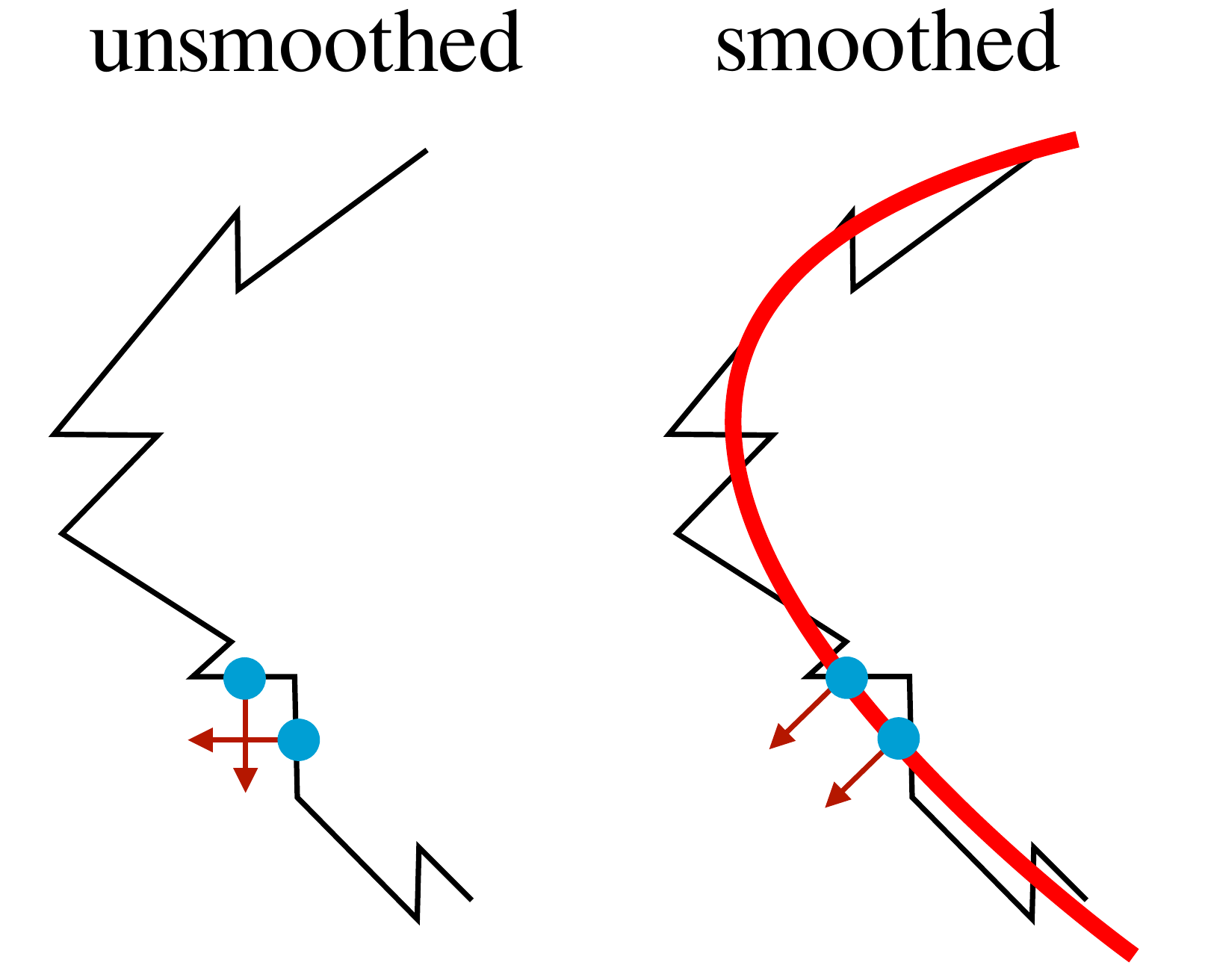}
  \end{center}
\end{wrapfigure}
Since $p_\beta(x)$ closely resembles a normal distribution with variance $\sigma = \log(2) \tfrac{\sqrt{2 \pi}}{\beta}$, $\beta$-smoothing can be understood as $N \to \infty$ limit of SmoothGrad $h(x)=\tfrac{1}{N}\sum_{i=1}^N \nabla g(x-\epsilon_i)$ where $\epsilon_i \sim g_\beta \approx \mathcal{N}(0,\sigma)$. We emphasize that the theorem only holds for a one-layer neural network, but for deeper networks we empirically observe that both lead to visually similar maps as they are considerably less noisy than the gradient map. The theorem therefore suggests that SmoothGrad can similarly be used to smooth the curvatures and can thereby make explanations more robust.\footnote{For explanation methods $h(x)$ other than gradient, SmoothGrad needs to be used in a slightly generalized form, i.e. $\tfrac{1}{N}\sum_{i=1}^N h(x-\epsilon_i)$.}

\textbf{Experiments:} Figure~\ref{fig:softplus} demonstrates that $\beta$-smoothing allows us to recover the orginal explanation map by lowering the value of the $\beta$ parameter. We stress that this works for all considered methods. We also note that the same effect can be observed using SmoothGrad by successively increasing the standard deviation $\sigma$ of the noise distribution. This further underlines the similarity between the two smoothing methods.

\begin{figure}[h!]
  \centering
  \includegraphics[width=.8\linewidth]{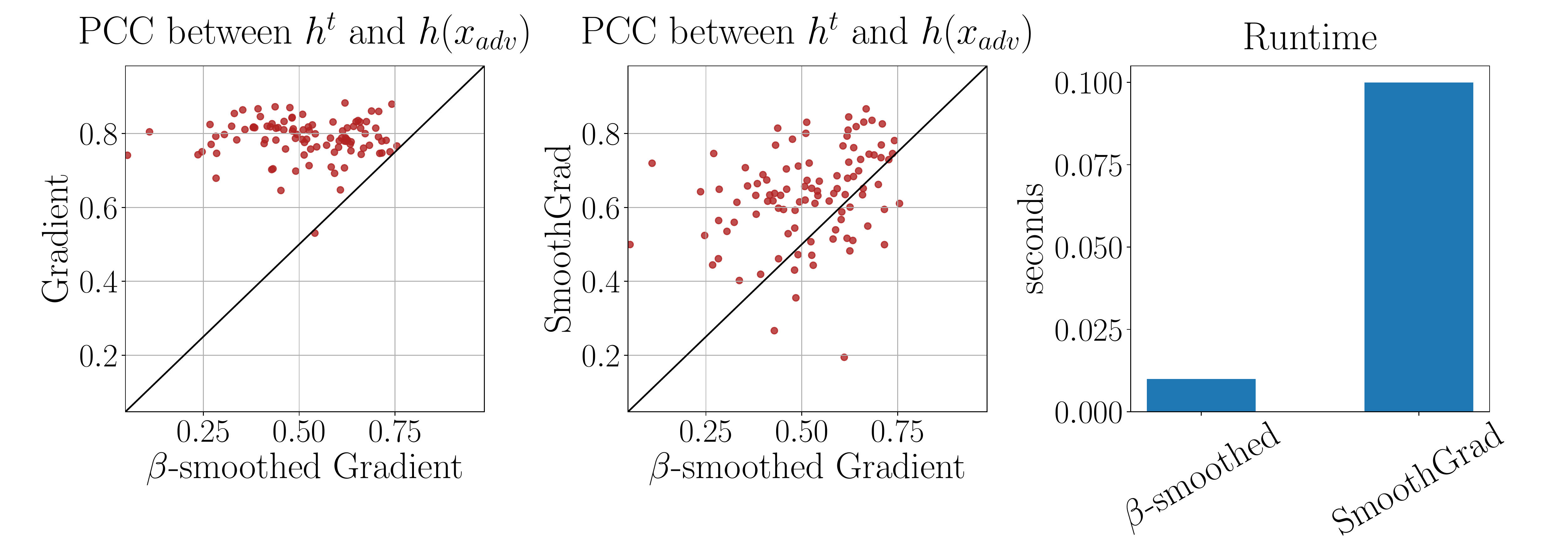}
   \caption{Left: markers are clearly left of the diagonal, i.e. explanations are more robust to manipulations when $\beta$-smoothing is used. Center: SmoothGrad has comparable results to $\beta$-smoothing, i.e. markers are distributed around the diagonal. Right:  $\beta$-smoothing has significantly lower computational cost than SmoothGrad.\label{fig:vanilla_soft_smooth}}
\end{figure}

If an attacker knew that smoothing was used to undo the manipulation, they could try to attack the smoothed method directly. However, both $\beta$-smoothing and SmoothGrad are substantially more robust than their non-smoothed counterparts, see Figure~\ref{fig:vanilla_soft_smooth}. It is important to note that $\beta$-smoothing achieves this at considerably lower computational cost: $\beta$-smoothing only requires a single forward and backward pass, while SmoothGrad requires as many as the number of noise samples (typically between 10 to 50).

We refer to Supplement~\ref{app:smoothing} for more details on these experiments.

\section{Conclusion}
Explanation methods have recently become increasingly  popular among practitioners. In this contribution we show that dedicated imperceptible manipulations of the input data can yield arbitrary and drastic changes of the explanation map. We demonstrate both qualitatively and quantitatively that explanation maps of many popular explanation methods can be arbitrarily manipulated. Crucially, this can be achieved while keeping the model’s output constant. A novel theoretical analysis reveals that in fact the large curvature of the network's decision function is one important culprit for this unexpected vulnerability. Using this theoretical insight, we can profoundly increase the resilience to manipulations by smoothing {\em only} the explanation process while leaving the model itself unchanged. 

Future work will investigate possibilities to  modify the training process of neural networks itself such that they can  become less vulnerable to manipulations of explanations. Another interesting future direction is to generalize our theoretical analysis from gradient to propagation-based methods. This seems particularly promising because our experiments strongly suggest that similar theoretical findings should also hold for these explanation methods.

\bibliography{references}
\bibliographystyle{unsrt}

\unhidefromtoc

\maketitle
\appendix

\tableofcontents

\section{Details on experiments}\label{app:details_experiments}
We provide a \emph{run\_attack.py} file in our reference implementation which allows one to produce manipulated images. 
The hyperparameter choices used in our experiments are summarized in Table~\ref{table:abssum}. We set $\beta_0=10$ and $\beta_e=100$ for beta growth (see section below for a description). The column 'factors' summarizes the weighting of the mean squared error of the heatmaps and the images respectively.
\begin{table}[h!]
\centering
\begin{tabular}{||c c c c||} 
 \hline
 method & iterations & lr & factors \\ [0.5ex] 
 \hline\hline
 Gradient & 1500 & $10^{-3}$ & $10^{11}$, $10^{6}$ \\
 Grad x Input & 1500 & $10^{-3}$ & $10^{11}$, $10^{6}$ \\
 IntGrad & 500 & $5 \times 10^{-3}$ & $10^{11}$, $10^{6}$ \\
 LRP & 1500 & $2 \times 10^{-4}$ & $10^{11}$, $10^{6}$ \\ 
 GBP & 1500 & $10^{-3}$ & $10^{11}$, $10^{6}$\\
 PA & 1500 & $2 \times 10^{-3}$ & $10^{11}$, $10^{6}$ \\
 [1ex] 
 \hline
\end{tabular}
\caption{Hyperparameters used in our analysis.}
\label{table:abssum}
\end{table}

The patterns for explanation method PA are trained on a subset of the ImageNet training set. The baseline $\bar{x}$ for explanation method IG was set to zero. To approximate the integral, we use $30$ steps for which we verified that the attributions approximately adds up to the score at the input.

\subsection{Beta growth}\label{app:beta_growth}
\begin{figure}[h]
  \centering
  \includegraphics[width=.8\linewidth]{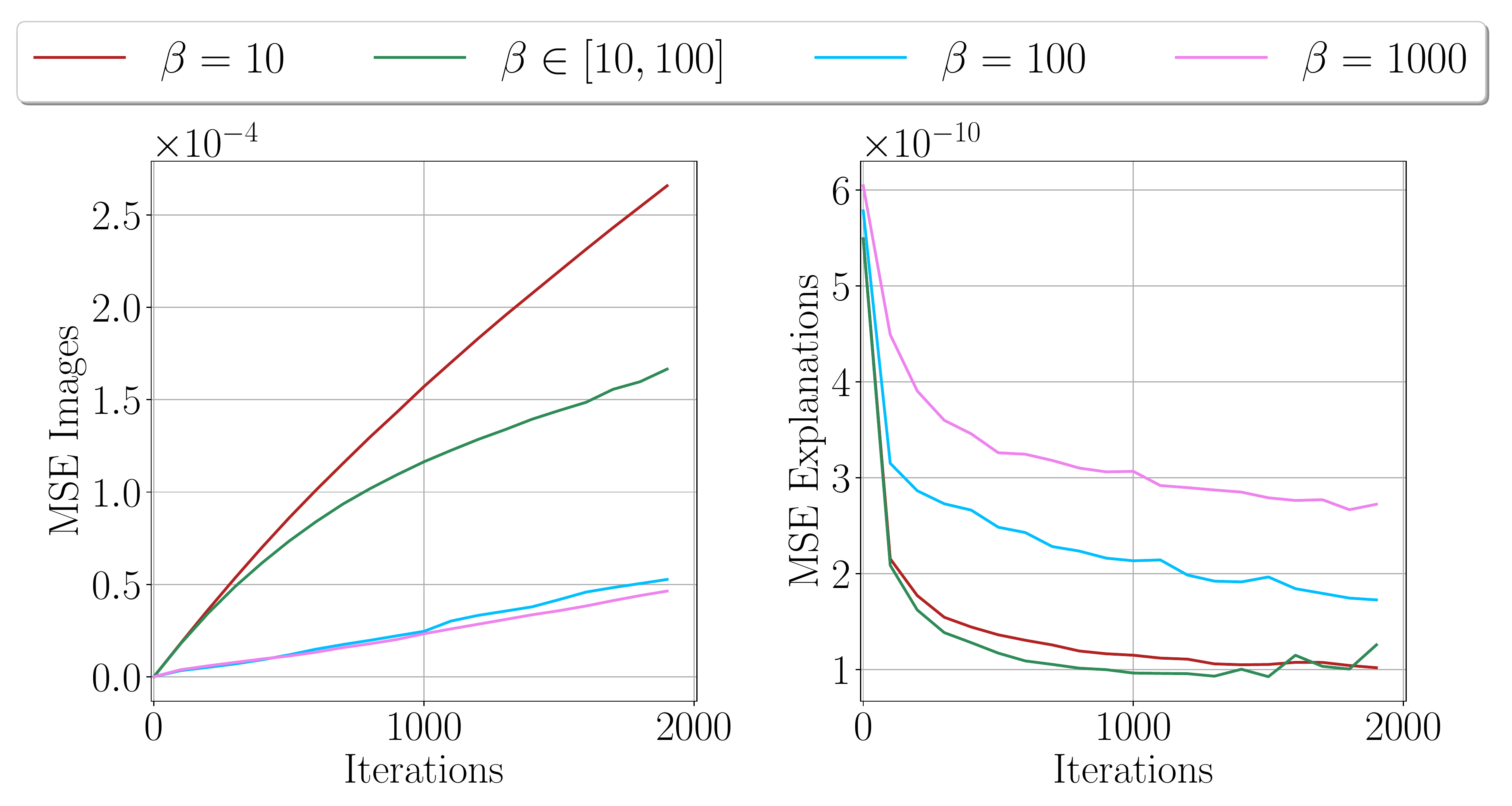}
  \caption{MSE between $x$ and $x_{adv}$ (left) and between $h^t$ and $h(x_{adv})$ (right) for various values for $\beta$.\label{fig:beta_growth}}
\end{figure}

In practise, we observe that we get slightly better results by increasing the value of $\beta$ of the softplus $sp(x)=\frac{1}{\beta}\ln{(1+e^{\beta x})}$ during training a start value $\beta_0$ to a final value $\beta_e$ using
\begin{align}
\beta(t) = \beta_0 \left(\frac{\beta_e}{\beta_0}\right)^{t/T} \label{eq:beta_growth}\,,
\end{align}
where $t$ is the current optimization step and $T$ denotes the total number of steps. Figure~\ref{fig:beta_growth} shows the MSE for images and explanation maps during training with and without $\beta$-growth. This strategy is however not essential for our results.

\begin{figure}[h]
  \centering
  \includegraphics[width=1.0\linewidth]{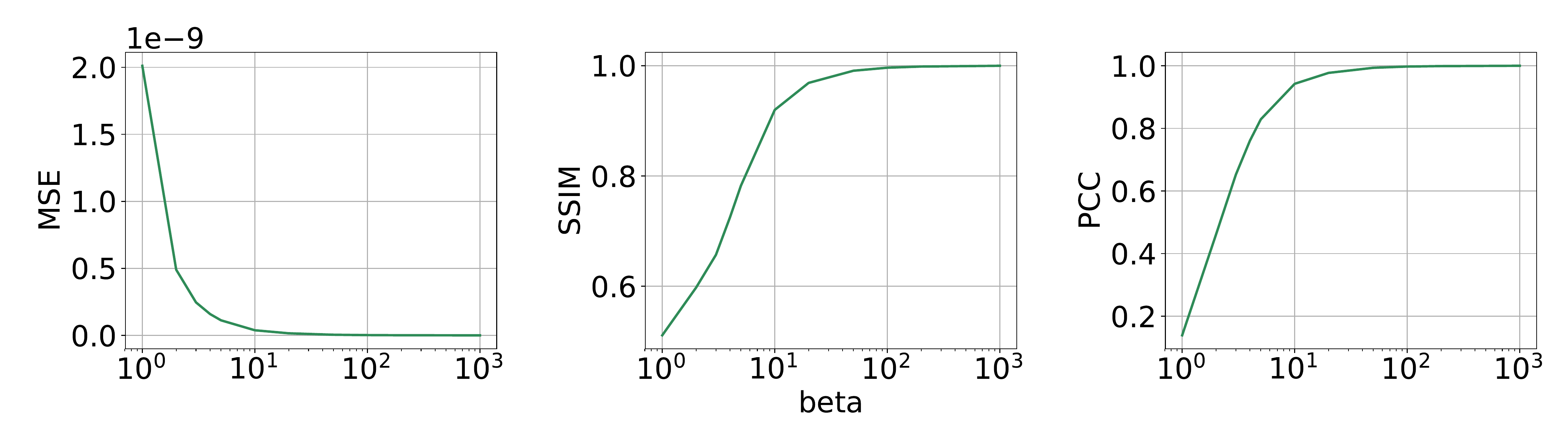}
  \caption{Error measures between the gradient explanation map produced with the original network and explanation maps produced with a network with softplus activation functions using various values for $\beta$.\label{fig:error_measures_beta_growth_gradient}}
\end{figure}

\begin{figure}[h!]
  \centering
  \includegraphics[width=1.0\linewidth]{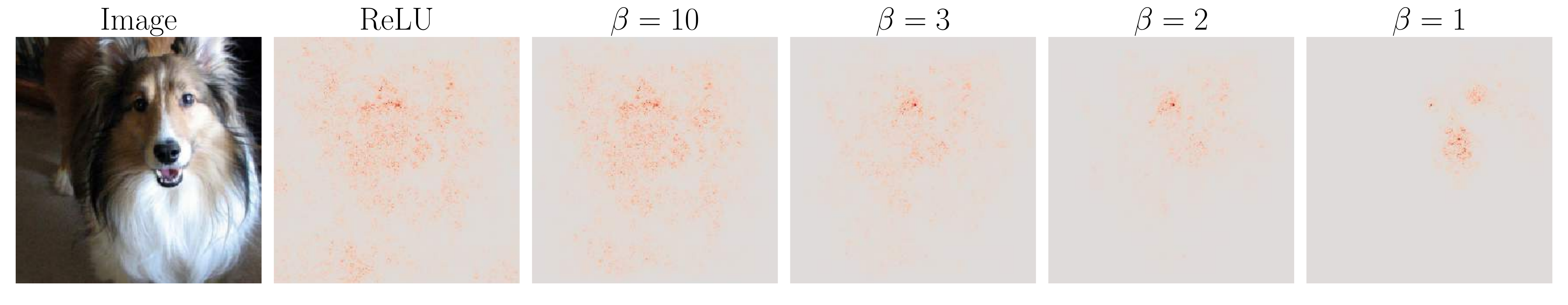}
  \caption{Gradient explanation map produced with the original network and a network with softplus activation functions using various values for $\beta$. \label{fig:heatmaps_beta_growth_gradient}}
\end{figure}

\begin{figure}[h!]
  \centering
  \includegraphics[width=1.0\linewidth]{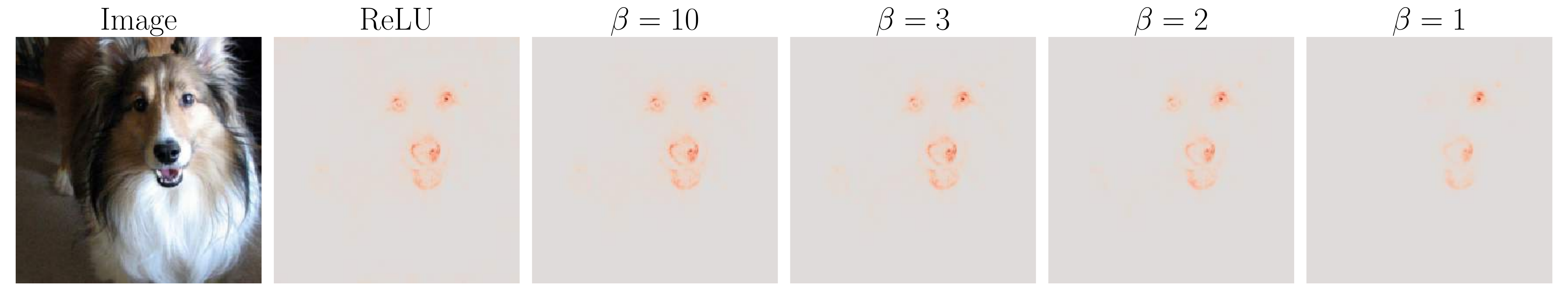}
  \caption{LRP explanation map produced with the original network and a network with softplus activation functions using various values for $\beta$. \label{fig:heatmaps_beta_growth_lrp}}
\end{figure}

We use beta growth for all methods except LRP for which we do not find any speed-up in the optimization as the LRP rules do not explicitly depend on the second derivative of the relu activations.
Figure~\ref{fig:error_measures_beta_growth_gradient} demonstrates that for large beta values the softplus networks approximate the relu network well. Figure~\ref{fig:heatmaps_beta_growth_gradient} and Figure~\ref{fig:heatmaps_beta_growth_lrp} show this for an example for the gradient and the LRP explanation method. We also note that for small beta the gradient explanation maps become more similar to LRP/GPB/PA explanation maps.

\section{Difference in network output}\label{app:diff_output}
\begin{figure}[h]
  \centering
  \includegraphics[width=.7\linewidth]{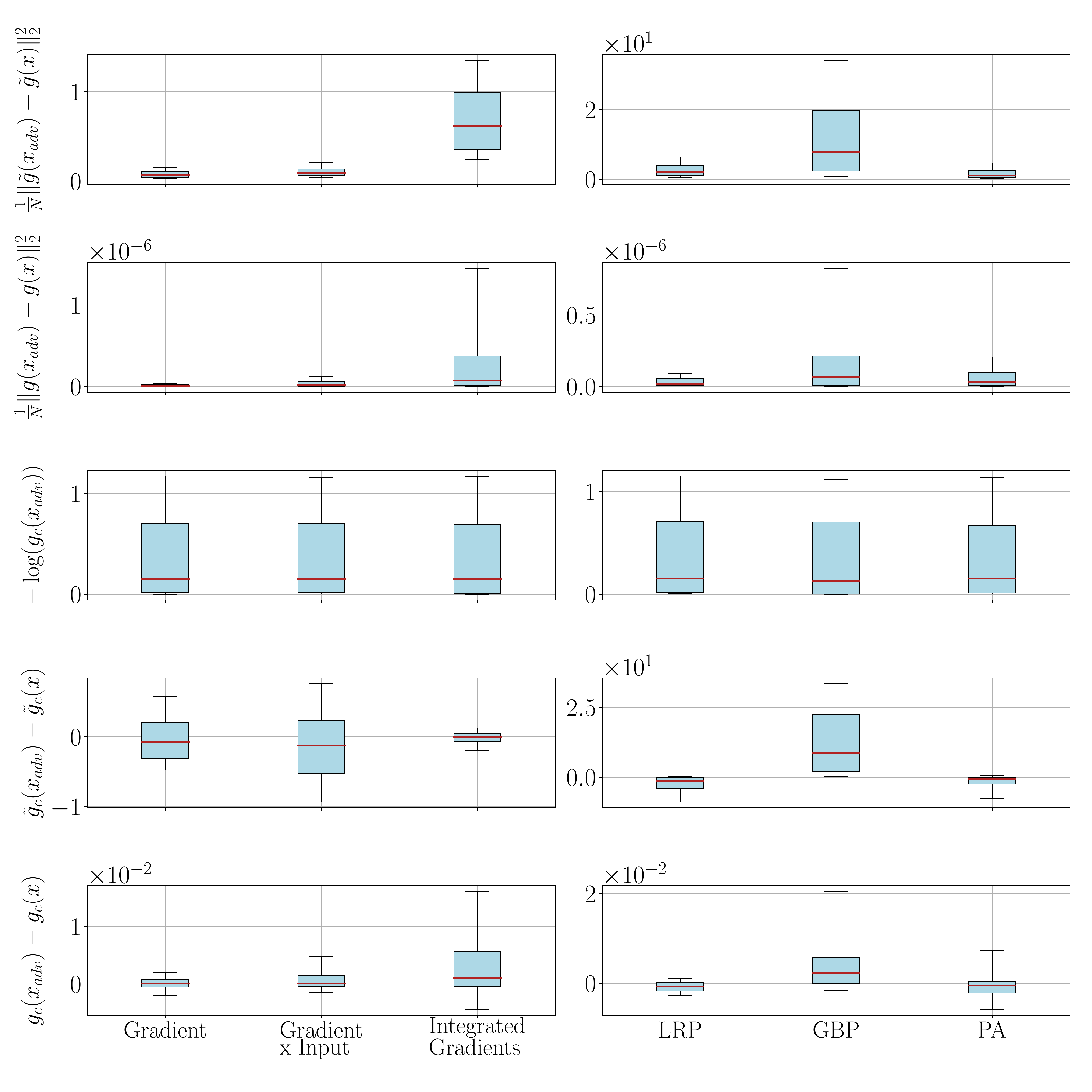}
  \caption{Error analysis of Network output. $\tilde{g}(x)$ denotes pre-activation of the last layer. $g(x)$ is the network output after applying the softmax function to the pre-activation $\tilde{g}(x)$.\label{fig:relu_net_diff_class_losses_cll}}
\end{figure}

Figure~\ref{fig:relu_net_diff_class_losses_cll} summarizes the change in the output of the network due to the manipulation. We note that all images have the same classification result as the orginals. Furthermore, we note that the change in confidence is small. Last but not least, norm of the vector of all class probabilities is also very small.

\FloatBarrier
\section{Generalization over architectures and data sets}\label{app:other_architectures_datasets}
Manipulable explanations are not only a property of the VGG-16 network. In this section, we show that our algorithm to manipulate explanations can also be applied to other architectures and data sets. For the experiments, we optimize the loss function given in the main text. 
We keep the pre-activation for all network architectures approximately constant, which also leads to approximately constant activation.

\subsection{Additional architectures}
\begin{figure}[h]
  \centering
  \includegraphics[width=.7\linewidth]{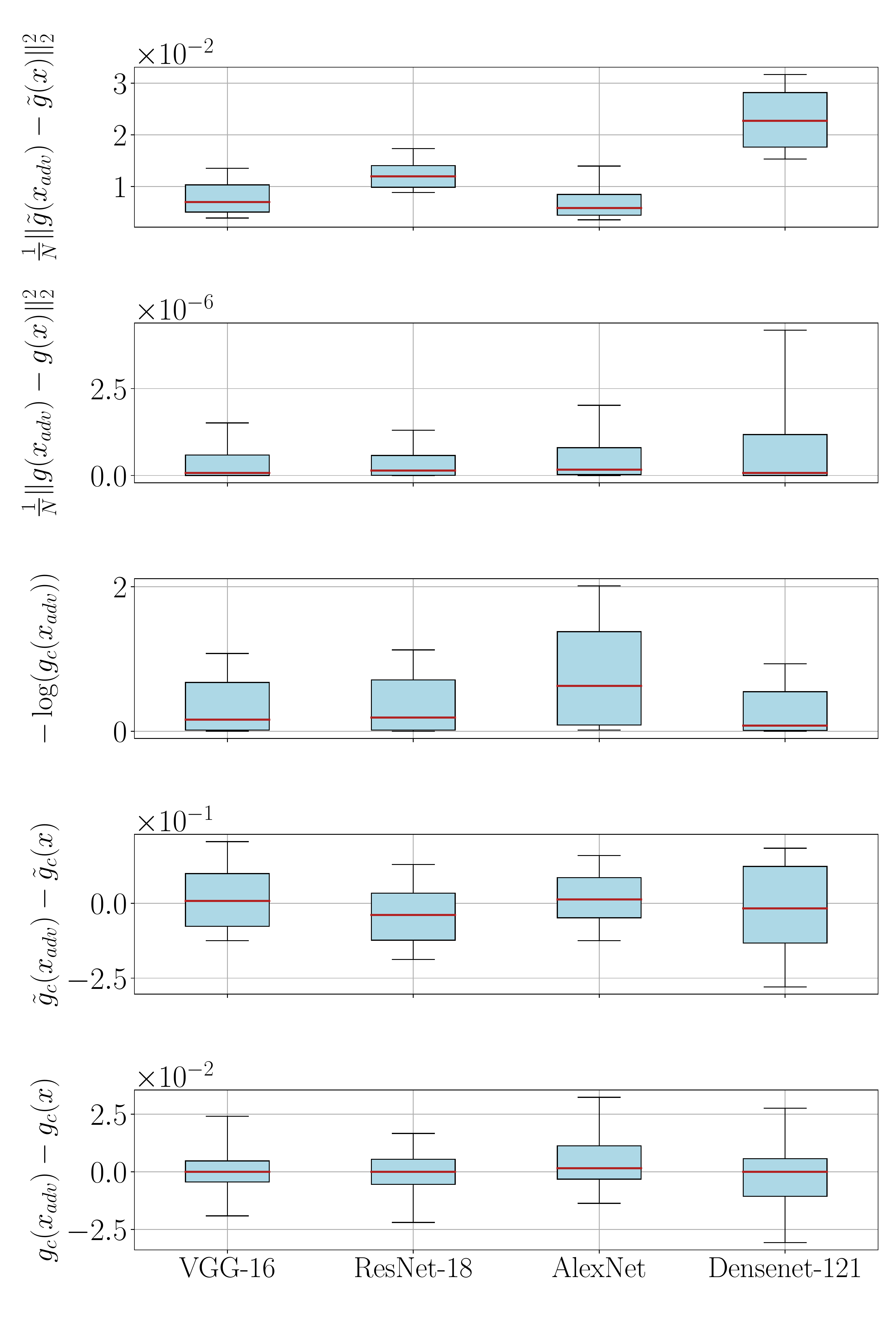}
  \caption{. Change in output for various architectures. \label{fig:all_nets_class_errors_cll}}
\end{figure}

\begin{figure}[h]
  \centering
  \includegraphics[width=1.0\linewidth]{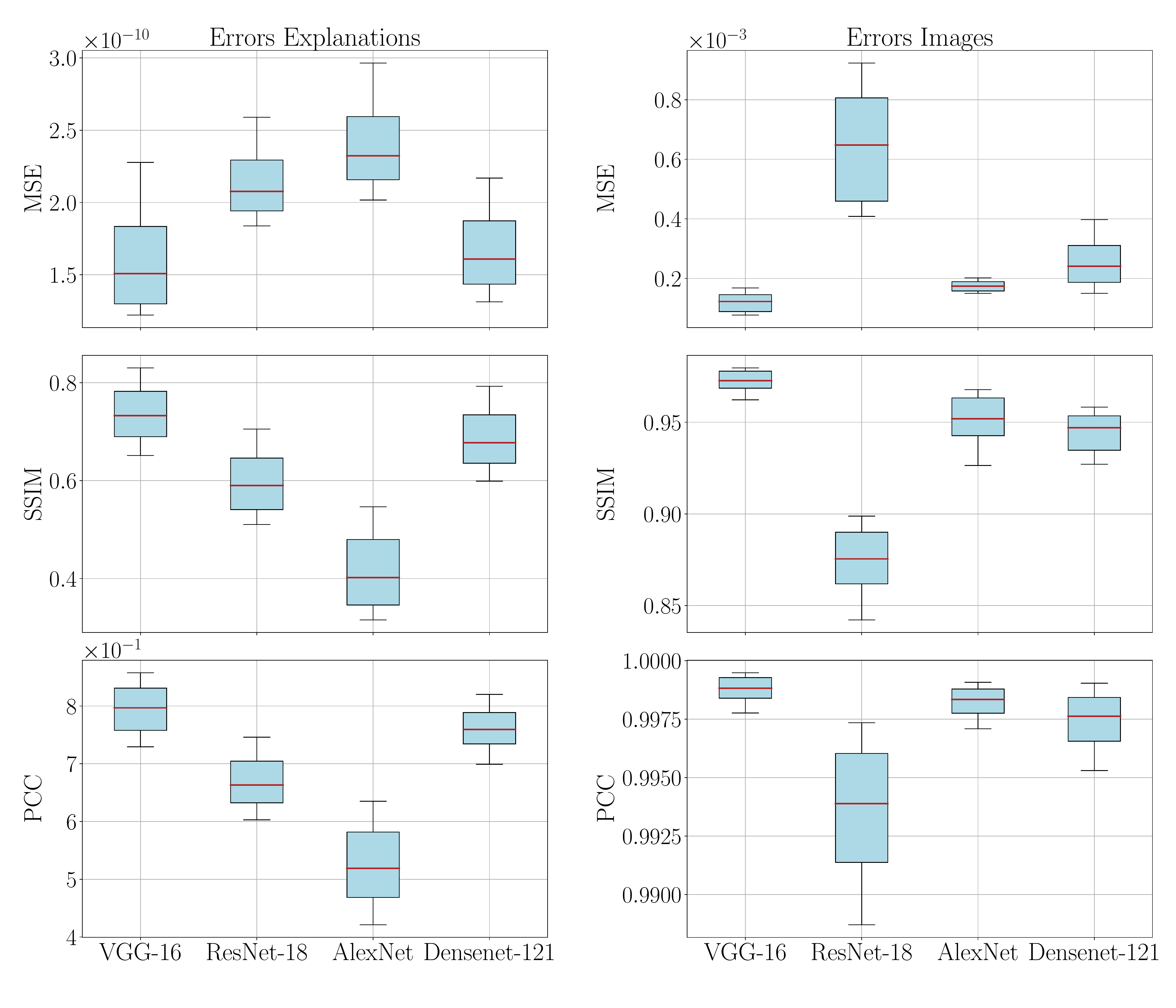}
  \caption{Similarity measures for gradient method for various architectures. \label{fig:all_nets_hm_image_error_cll}}
\end{figure}

In addition to the VGG architecture we also analyzed the explanation's susceptibility to manipulations for the AlexNet, Densenet and ResNet architectures. The hyperparameter choices used in our experiments are summarized in Table~\ref{tab:allnets}. We set $\beta_0=10$ and $\beta_e=100$ for beta growth. Only for Densenet we set $\beta_0=30$ and $\beta_e=300$ as for smaller beta values the explanation map produced with softplus does not resemble the explanation map produced with relu. Figure~\ref{fig:all_nets_hm_image_error_cll} and \ref{fig:all_nets_class_errors_cll} show that the similarity measures are comparable for all network architectures for the gradient method. 

\begin{figure}[h]
  \centering
  \includegraphics[width=1.0\linewidth]{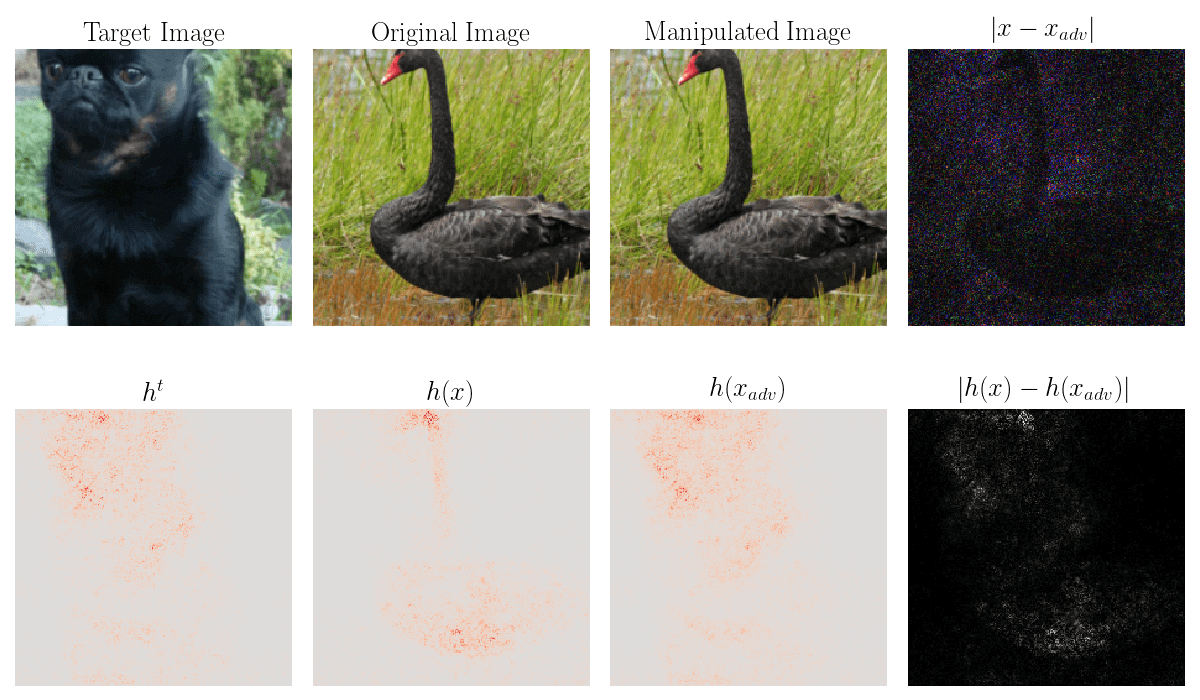}
  \caption{Gradient explanation maps produced with VGG-16 model. \label{fig:overview_vgg}}
\end{figure}

\begin{figure}[h]
  \centering
  \includegraphics[width=1.0\linewidth]{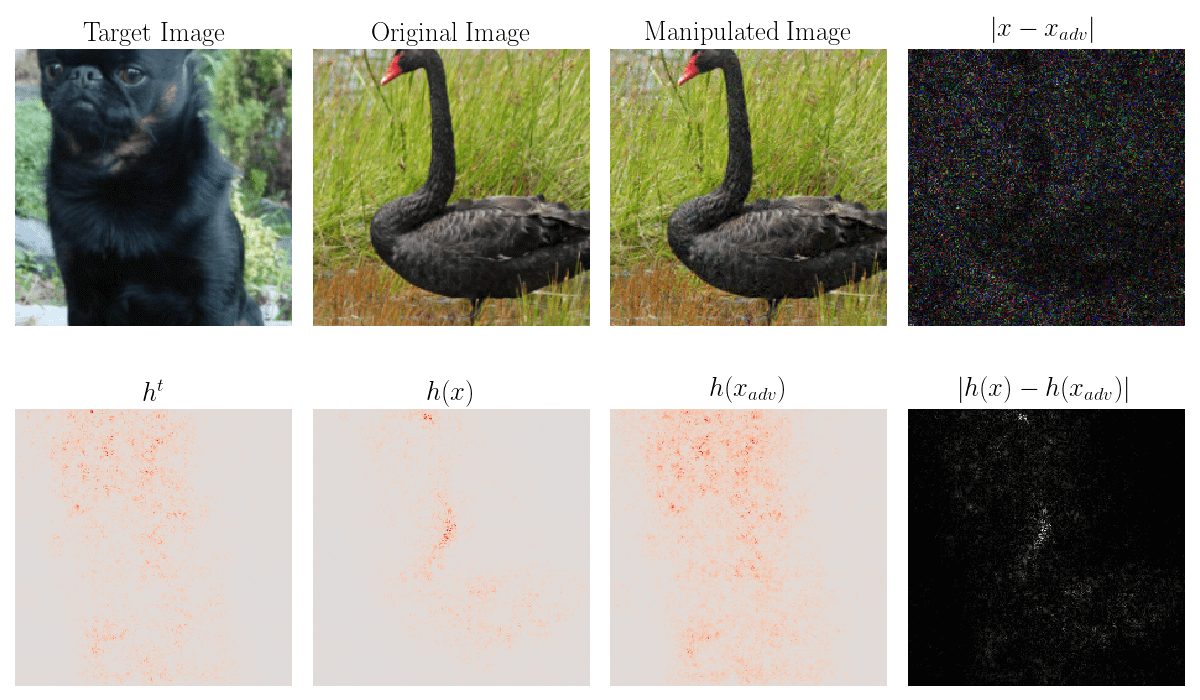}
  \caption{Gradient explanation maps produced with ResNet-18 model. \label{fig:overview_resnet}}
\end{figure}

\begin{figure}[h]
  \centering
  \includegraphics[width=1.0\linewidth]{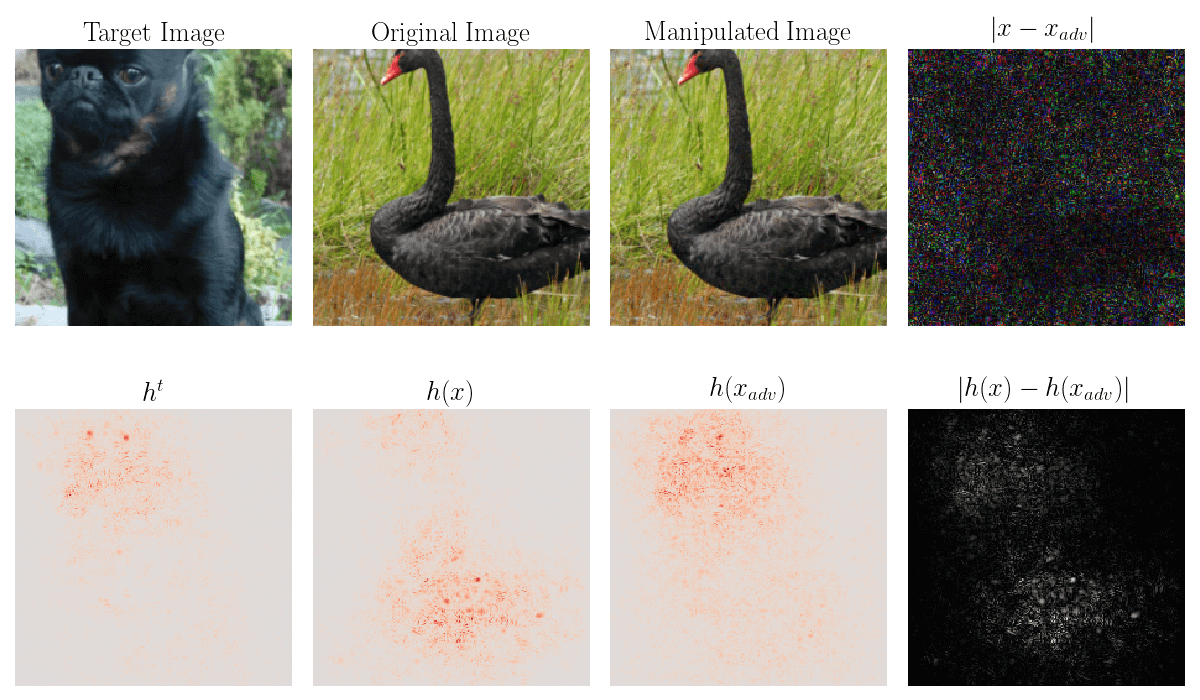}
  \caption{Gradient explanation maps produced with AlexNet model. \label{fig:overview_alexnet}}
\end{figure}

\begin{figure}[h]
  \centering
  \includegraphics[width=1.0\linewidth]{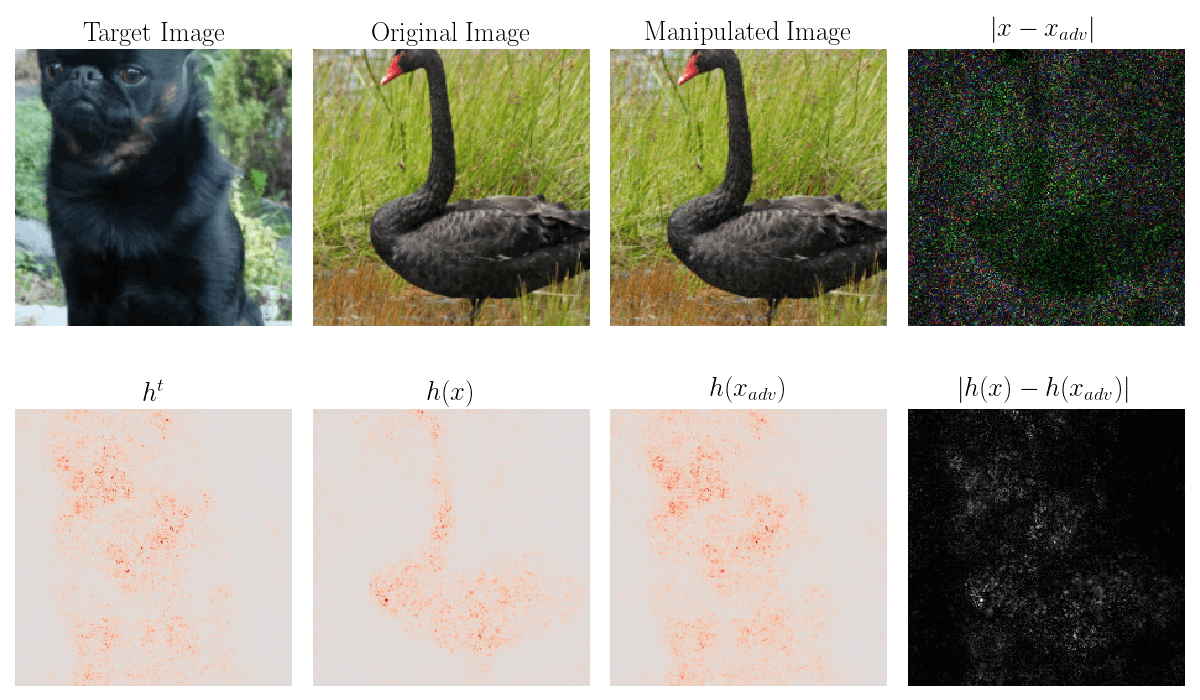}
  \caption{Gradient explanation maps produced with Densenet-121 model. \label{fig:overview_densenet}}
\end{figure}

Figure~\ref{fig:overview_vgg}, \ref{fig:overview_alexnet}, \ref{fig:overview_densenet} and \ref{fig:overview_resnet} show one example image for each architecture. 

\begin{table}[h!]
\centering
\begin{tabular}{||c c c c||} 
 \hline
 network & iterations & lr & factors \\ [0.5ex] 
 \hline\hline
 VGG16 & 1500 & $10^{-3}$ & 1e11, 10 \\ 
 AlexNet & 4000 & $10^{-3}$ & 1e11, 10 \\
 Densenet-121 & 2000 & $5\times10^{-4}$ & 1e11, 10 \\
 ResNet-18 & 2000 & $10^{-3}$ & 1e11, 10 \\
 [1ex] 
 \hline
\end{tabular}
\caption{Hyperparameters used in our analysis for all networks.}
\label{tab:allnets}
\end{table}


\subsection{Additional datasets}
\begin{figure}[h!]
\centering
\begin{subfigure}{.49\textwidth}
  \includegraphics[width=1\linewidth]{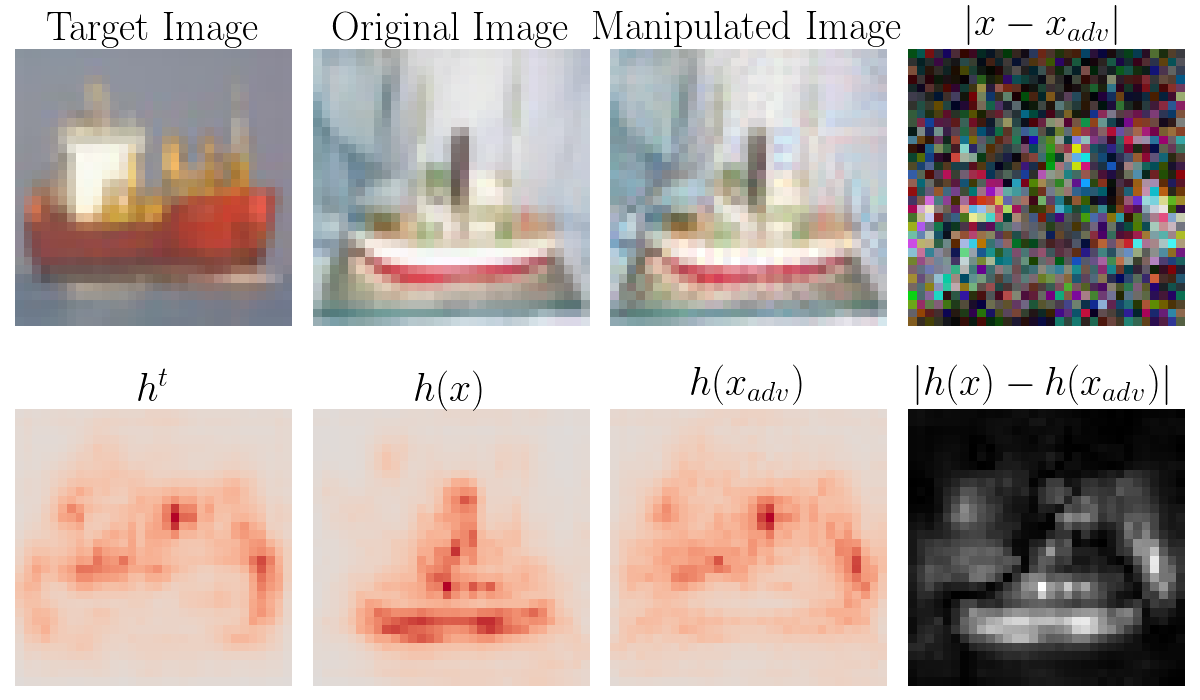}
\end{subfigure}%
\hfill
\begin{subfigure}{.49\textwidth}
  \includegraphics[width=1\linewidth]{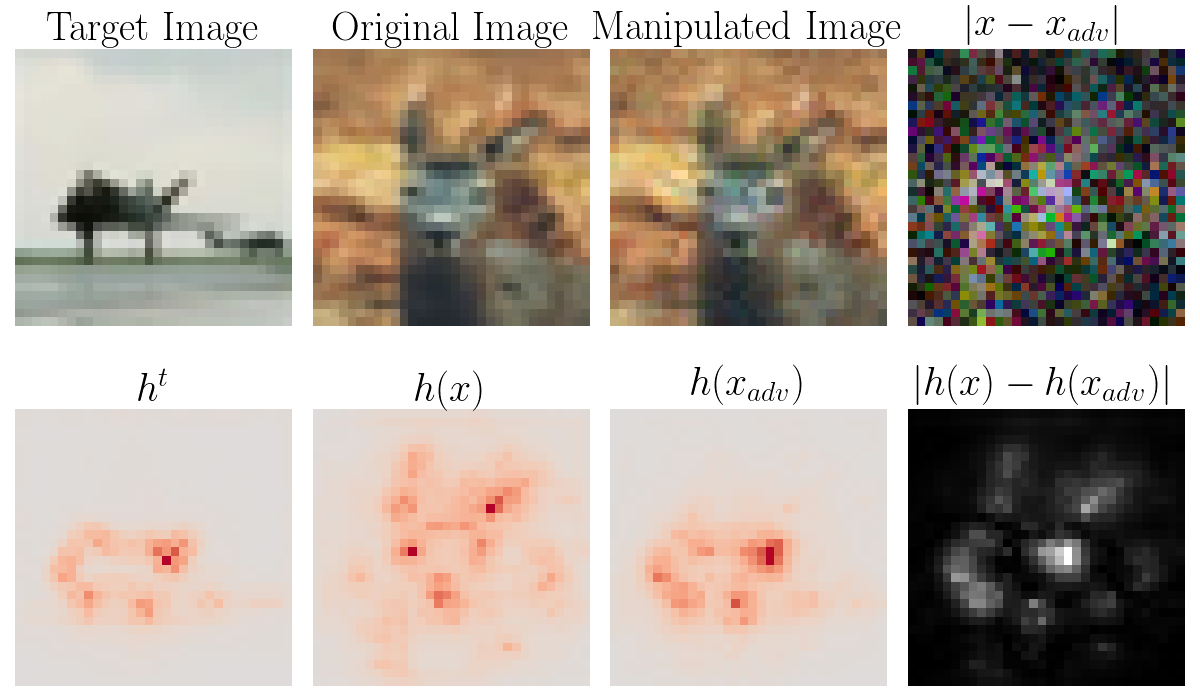}
 \end{subfigure}
 \caption{LRP Method on CIFAR-10 dataset}
 \label{fig:cifar}
\end{figure}

We trained the VGG-16 architecture on the CIFAR-10 dataset\footnote{code for training VGG on CIFAR-10 from  \url{https://github.com/chengyangfu/pytorch-vgg-cifar10}}. The test accuracy is approximately $92\%$. We then used our algorithm to manipulate the explanations for the LRP method. The hyperparameters are summarized in Table~\ref{table:cifar}. Two example images can be seen in Figure~\ref{fig:cifar}.
\begin{table}[h!]
\centering
\begin{tabular}{||c c c c||} 
 \hline
 method & iterations & lr & factors \\ [0.5ex] 
 \hline\hline
 LRP & 1500 & $2 \times 10^{-4}$ & $10^{7}$, $10^{2}$ \\ 
 [1ex] 
 \hline
\end{tabular}
\caption{Hyperparameters used in our analysis for the CIFAR-10 Dataset.}
\label{table:cifar}
\end{table}

\FloatBarrier
\section{Smoothing explanation methods}\label{app:smoothing}
One can achieve a smoothing effect when substituting the relu activations for softplus$_{\beta}$ activations and then applying the usual rules for the different explanation methods.

\begin{figure}[h!]
\centering
\begin{subfigure}{.17\textwidth}
  \includegraphics[width=1\linewidth]{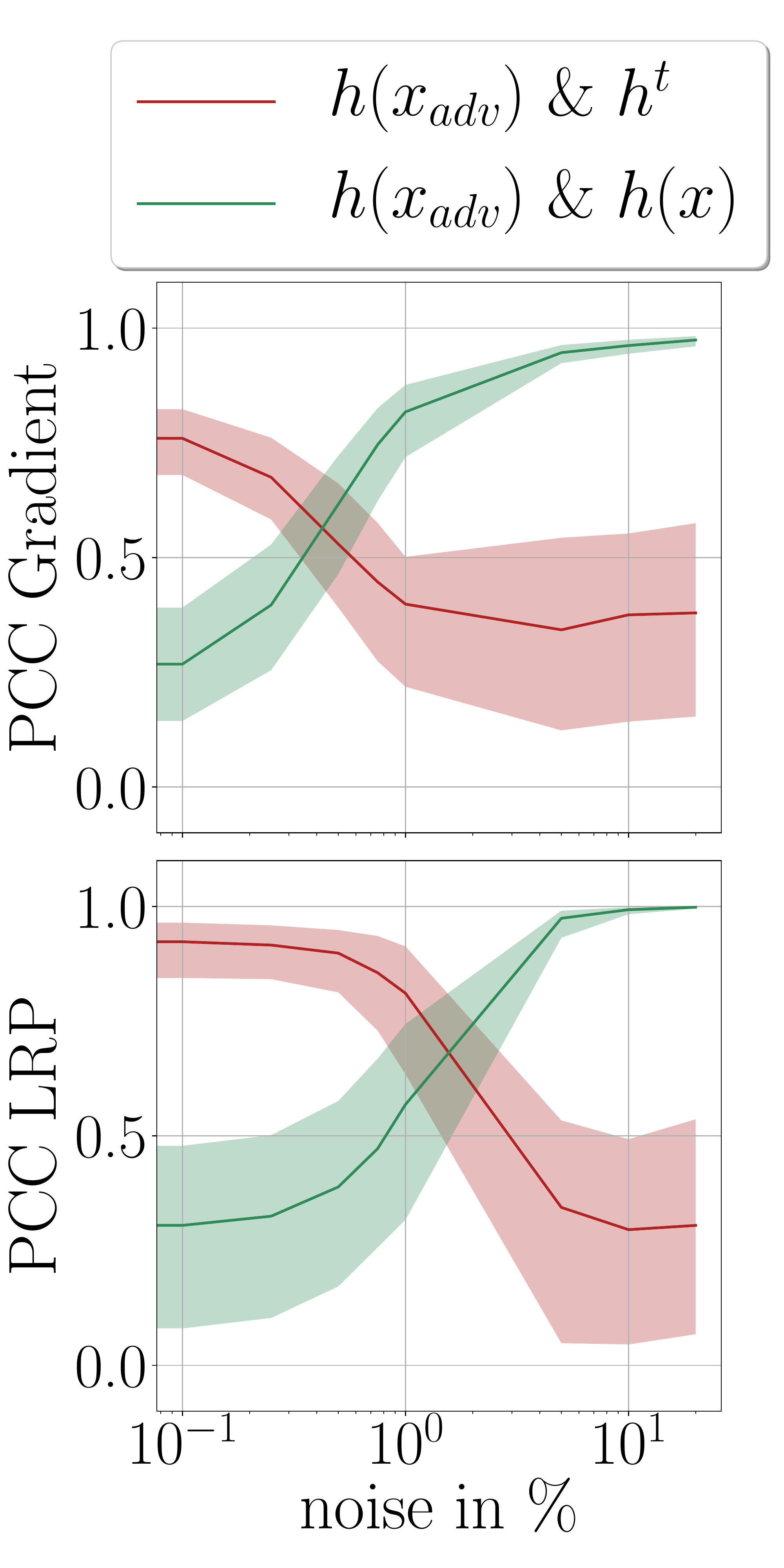}
\end{subfigure}%
\hfill
\begin{subfigure}{.4\textwidth}
  \includegraphics[width=1\linewidth]{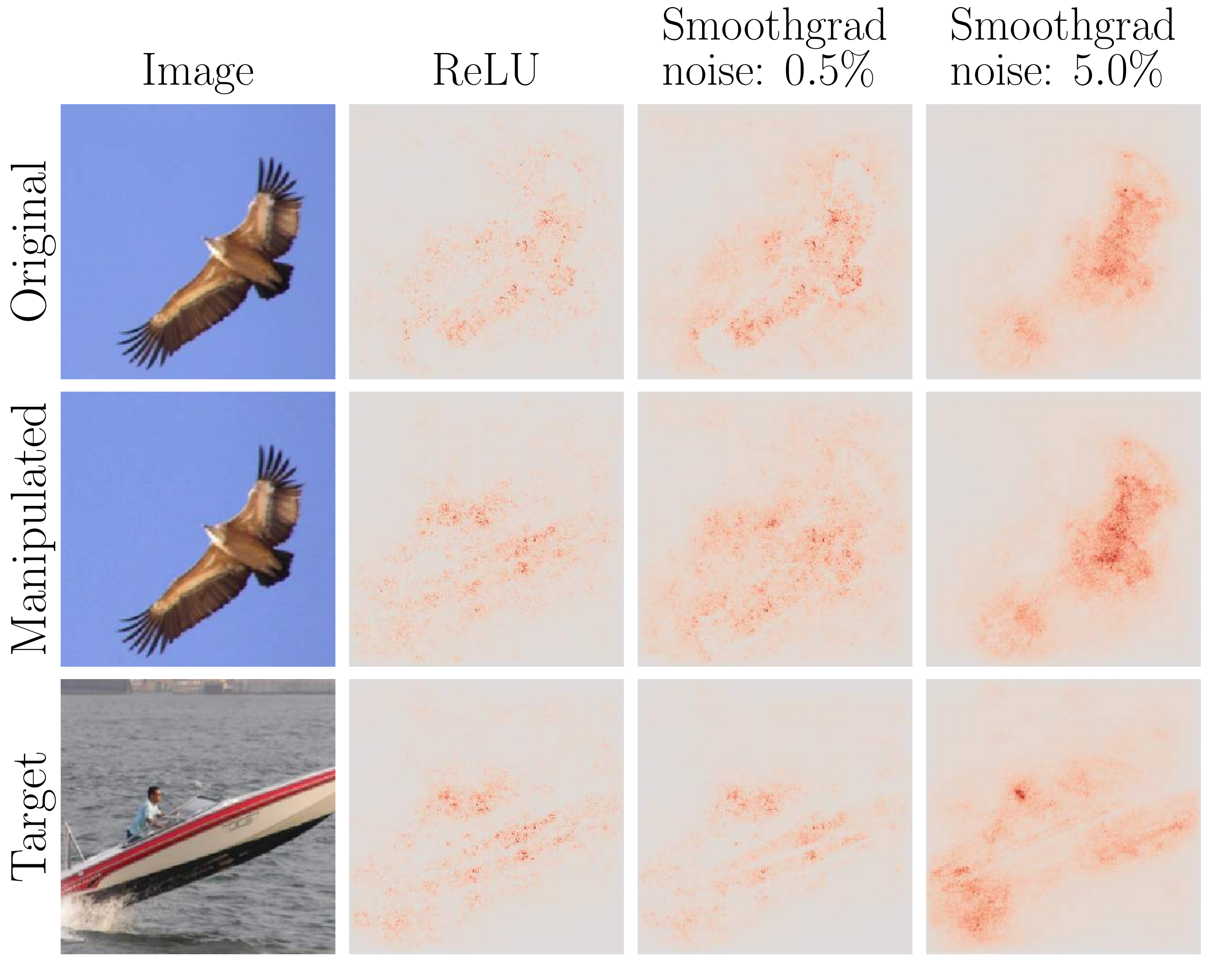}
 \end{subfigure}
 \begin{subfigure}{.4\textwidth}
  \includegraphics[width=1\linewidth]{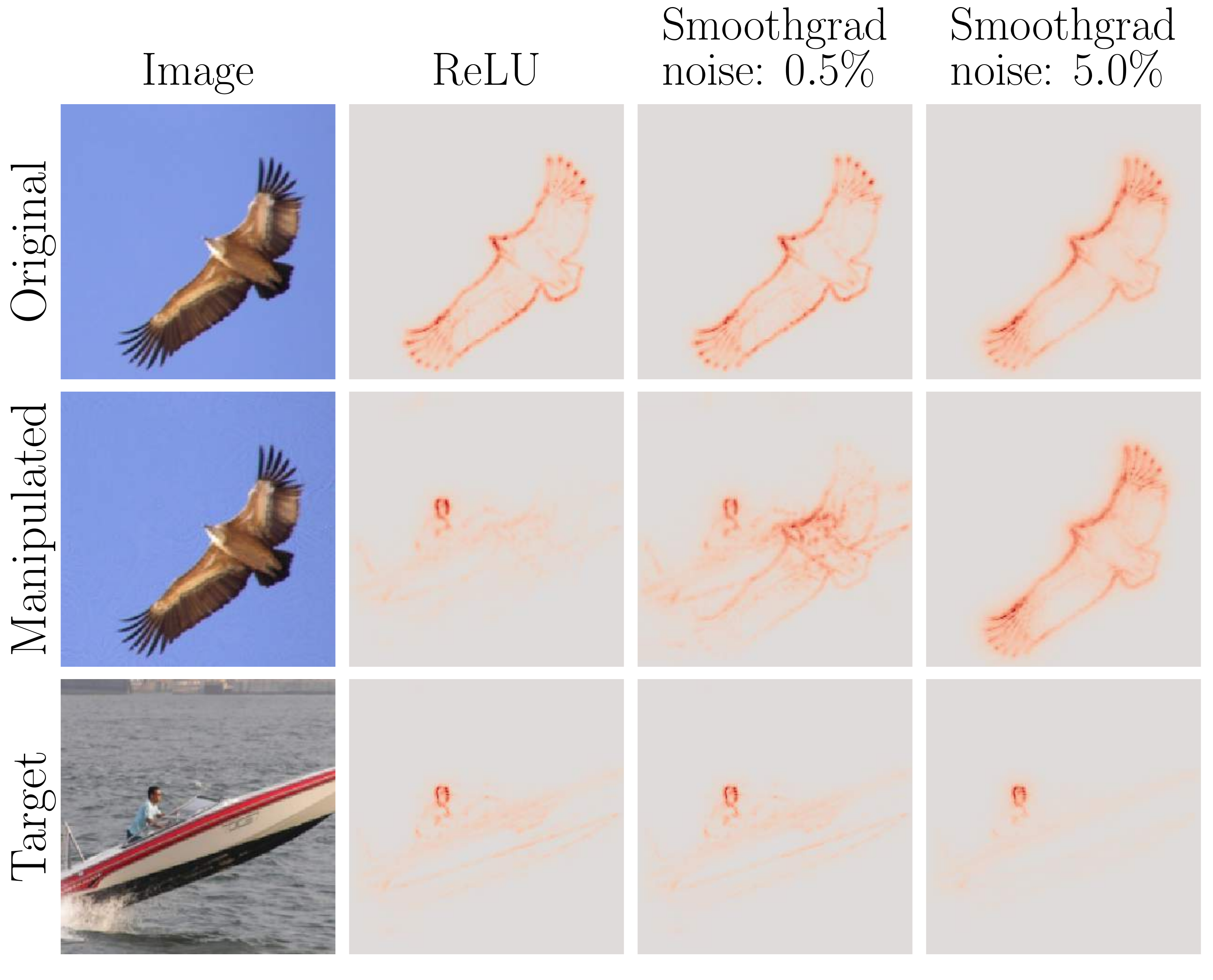}
 \end{subfigure}
 \caption{Recovering the original explanation map with SmoothGrad. Left: $\beta$ dependence for the correlations of the manipulated explanation (here Gradient and LRP) with the target and original explanation. Line denotes median, $10^{th}$ and $90^{th}$ percentile are shown in semitransparent colour. Center and Right: network input and the respective explanation maps as $\beta$ is lowered for Gradient (center) and LRP (right). }
 \label{fig:smoothgrad}
\end{figure}

A smoothing effect can also be achieved by applying the smoothgrad explanation method, see Figure~\ref{fig:smoothgrad}. That is adding random perturbation to the image and then averaging over the resulting explanation maps. We average over 10 perturbed images with different values for the standard deviation $\sigma$ of the Gaussian noise. The noise level $n$ is related to $\sigma$ as $\sigma= n\cdot(x_{max}-x_{min})$, where $x_{max}$ and $x_{min}$ are the maximum and minimum values the input image can have.

\begin{figure}[h]
  \centering
  \includegraphics[width=.8\linewidth]{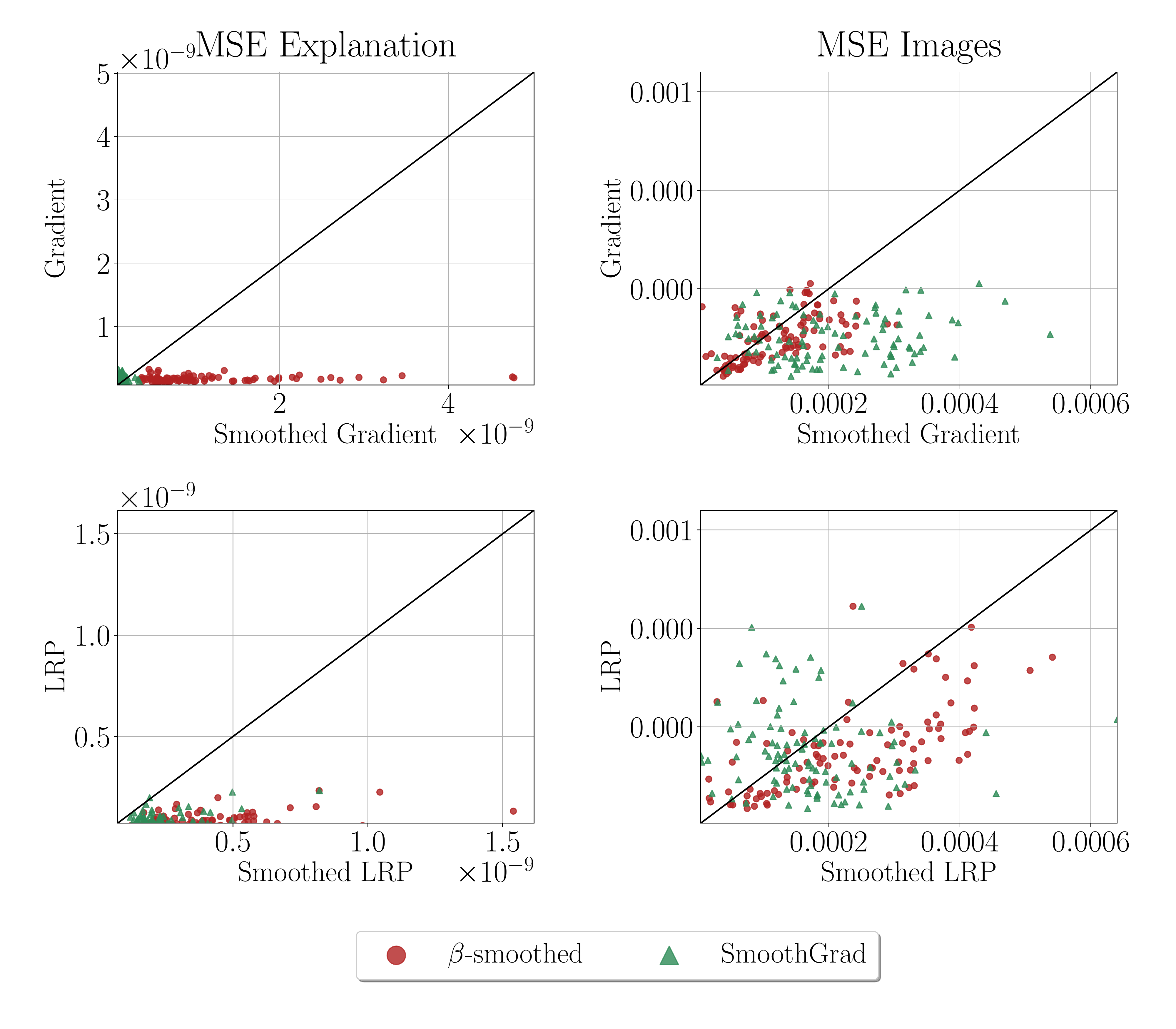}
\caption{Left: Similarities between explanations. Markers are mostly right of the diagonal, i.e. the MSE for the smoothed explanations is higher than for the unsmoothed explanations which means the manipulated smoothed explanation map does not closely resemble the target $h^t$.  Right: Similarities between Images. The MSE for the smoothed methods is higher (right of the diagonal) or comparable (on the diagonal), i.e. bigger or comparable perturbations in the manipulated Images when using smoothed explanation methods.\label{fig:soft_smooth_vanilla_grad_lrp_MSE}}
\end{figure}

\begin{figure}[h]
  \centering
  \includegraphics[width=.8\linewidth]{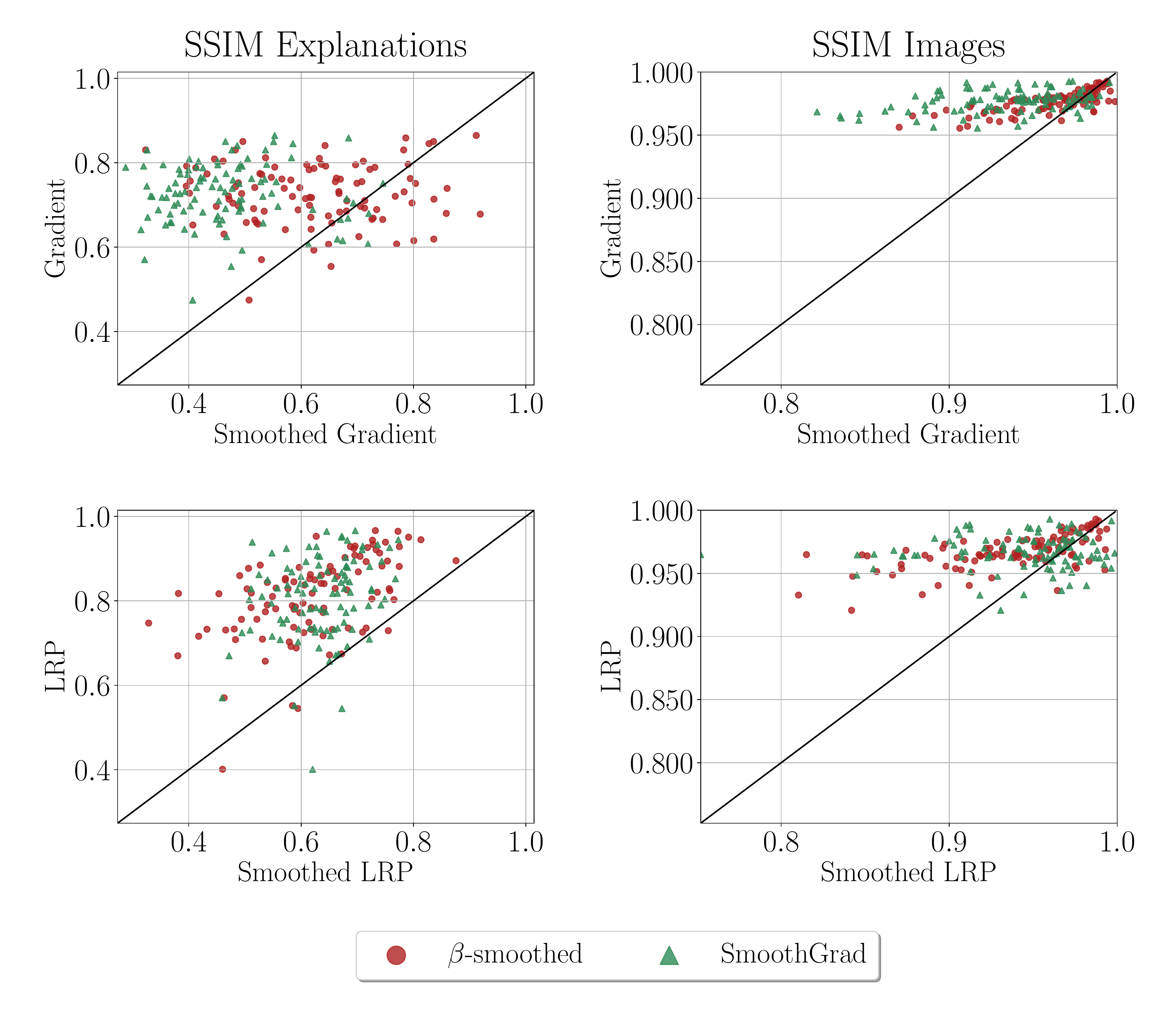}
\caption{Left: Similarities between explanations. Markers are mostly left of the diagonal, i.e. the SSIM for the smoothed explanations is lower than for the unsmoothed explanations which means the manipulated smoothed explanation map does not closely resemble the target $h^t$. Right: Similarities between Images. The SSIM for the smoothed methods is lower (left of the diagonal) or comparable (on the diagonal), i.e. bigger or comparable perturbations in the manipulated Images when using smoothed explanation methods.\label{fig:soft_smooth_vanilla_grad_lrp_SSIM}}
\end{figure}

\begin{figure}[h]
  \centering
  \includegraphics[width=.8\linewidth]{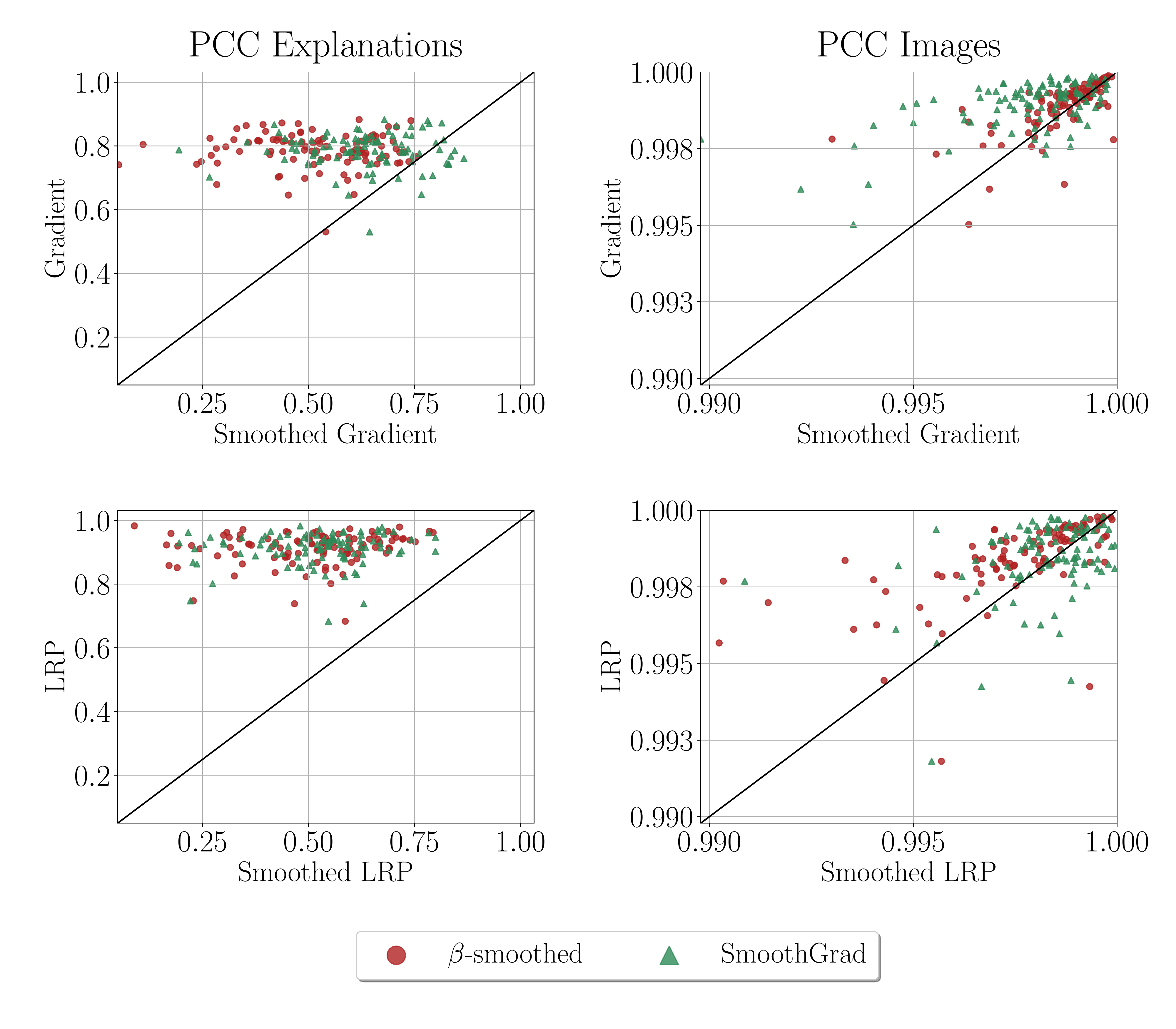}
\caption{Left: Similarities between explanations. Markers are mostly left of the diagonal, i.e. the PCC for the smoothed explanations is lower than for the unsmoothed explanations which means that the manipulated smoothed explanation map does not closely resemble the target $h^t$. Right: Similarities between Images. The PCC for the smoothed methods is lower (left of the diagonal) or comparable (on the diagonal), i.e. bigger or comparable perturbations in the manipulated Images when using smoothed explanation methods.\label{fig:soft_smooth_vanilla_grad_lrp_PCC}}
\end{figure}

The $\beta$-smoothing or SmoothGrad explanation maps are more robust with respect to manipulations.  Figure~\ref{fig:soft_smooth_vanilla_grad_lrp_MSE},~\ref{fig:soft_smooth_vanilla_grad_lrp_SSIM} and~\ref{fig:soft_smooth_vanilla_grad_lrp_PCC} show results (MSE, SSIM and PCC) for 100 targeted attacks on the original explanation, the SmoothGrad explanation and the $\beta$-smoothed explanation for explanation methods Gradient and LRP. 


For manipulation of SmoothGrad we use beta growth with $\beta_0=10$ and $\beta_e=100$. For manipulation of $\beta$-Smoothing we set $\beta=0.8$ for all runs. The hyperparameters for SmoothGrad and $\beta$-Smoothing are summarized in Table~\ref{table:smoothgrad_param} and Table~\ref{table:softplus_param}.

\begin{table}[h!]
\centering
\begin{tabular}{||c c c c||} 
 \hline
 method & iterations & lr & factors \\ [0.5ex] 
 \hline\hline
 Gradient & 1500 & $3 \times10^{-3}$ & $10^{11}$, $10^{6}$ \\
 LRP & 1500 & $3 \times 10^{-4}$ & $10^{11}$, $10^{6}$ \\
 [1ex] 
 \hline
\end{tabular}
\caption{Hyperparameters used in our analysis for SmoothGrad.}
\label{table:smoothgrad_param}
\end{table}

\begin{table}[h!]
\centering
\begin{tabular}{||c c c c||} 
 \hline
 method & iterations & lr & factors \\ [0.5ex] 
 \hline\hline
 Gradient & 500 & $2.5 \times10^{-4}$ & $10^{11}$, $10^{6}$ \\
 Grad x Input & 500 & $2.5 \times10^{-4}$ & $10^{11}$, $10^{6}$ \\
 IntGrad & 200 & $2.5 \times 10^{-3}$ & $10^{11}$, $10^{6}$ \\
 LRP & 1500 & $2.0 \times 10^{-4}$ & $10^{11}$, $10^{6}$ \\ 
 GBP & 500 & $5.0 \times10^{-4}$ & $10^{11}$, $10^{6}$\\
 PA & 500 & $5.0 \times10^{-4}$ & $10^{11}$, $10^{6}$ \\
 [1ex] 
 \hline
\end{tabular}
\caption{Hyperparameters used in our analysis for $\beta$-smoothing.}
\label{table:softplus_param}
\end{table}

\begin{figure}[h]
  \centering
  \includegraphics[width=1\linewidth]{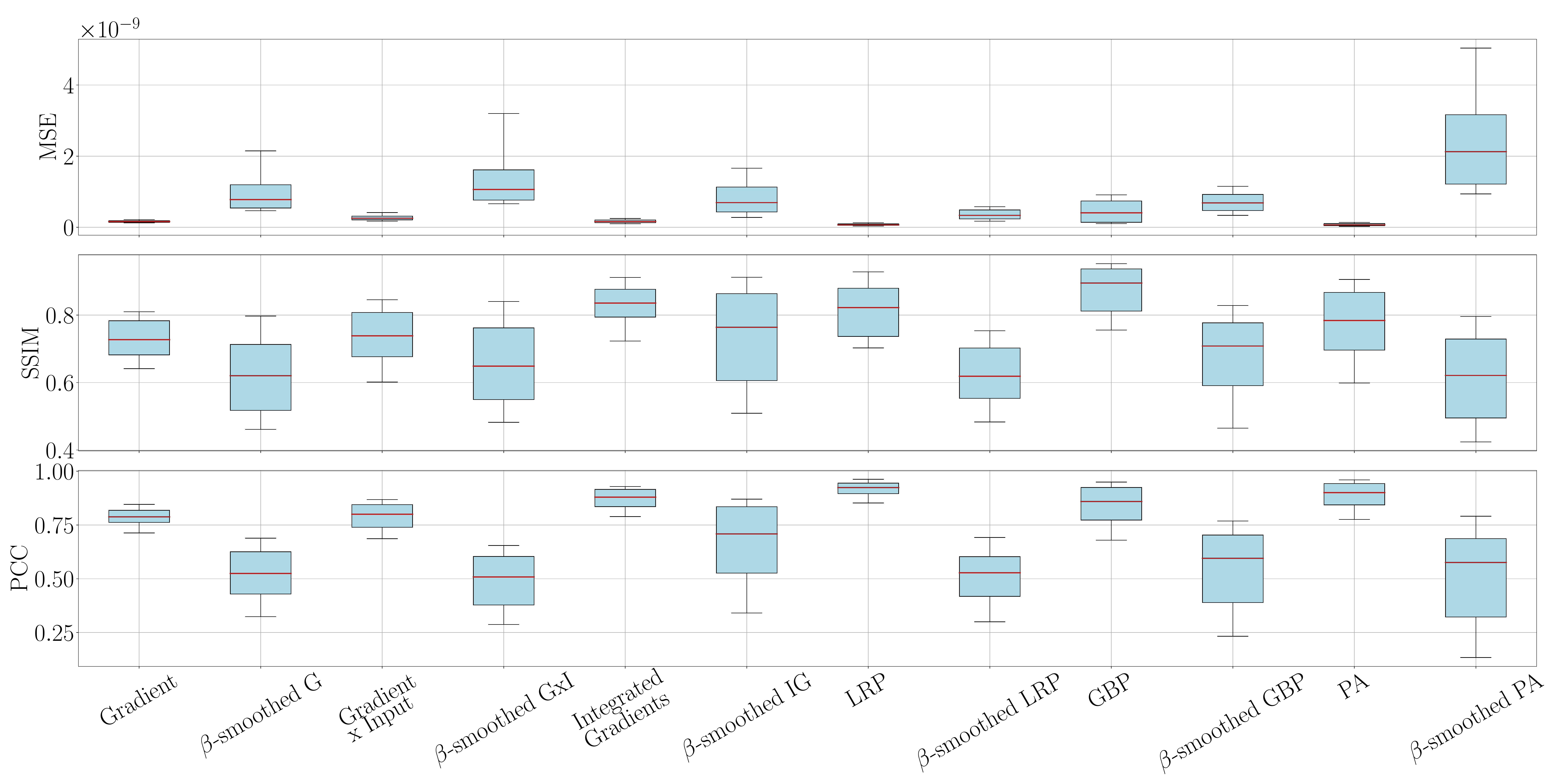}
  \caption{Comparison of Similarities of Explanation Maps for the original Explanation Methods and the $\beta$-smoothed Explanation Methods. Targeted attacks do not work very well on $\beta$-smoothed explanations, i.e. MSE is higher and SSIM and PCC are lower for the $\beta$-smoothed explanations than for the original explanations.\label{fig:vanilla_vs_soft_hm}}
\end{figure}

\begin{figure}[h]
  \centering
  \includegraphics[width=1\linewidth]{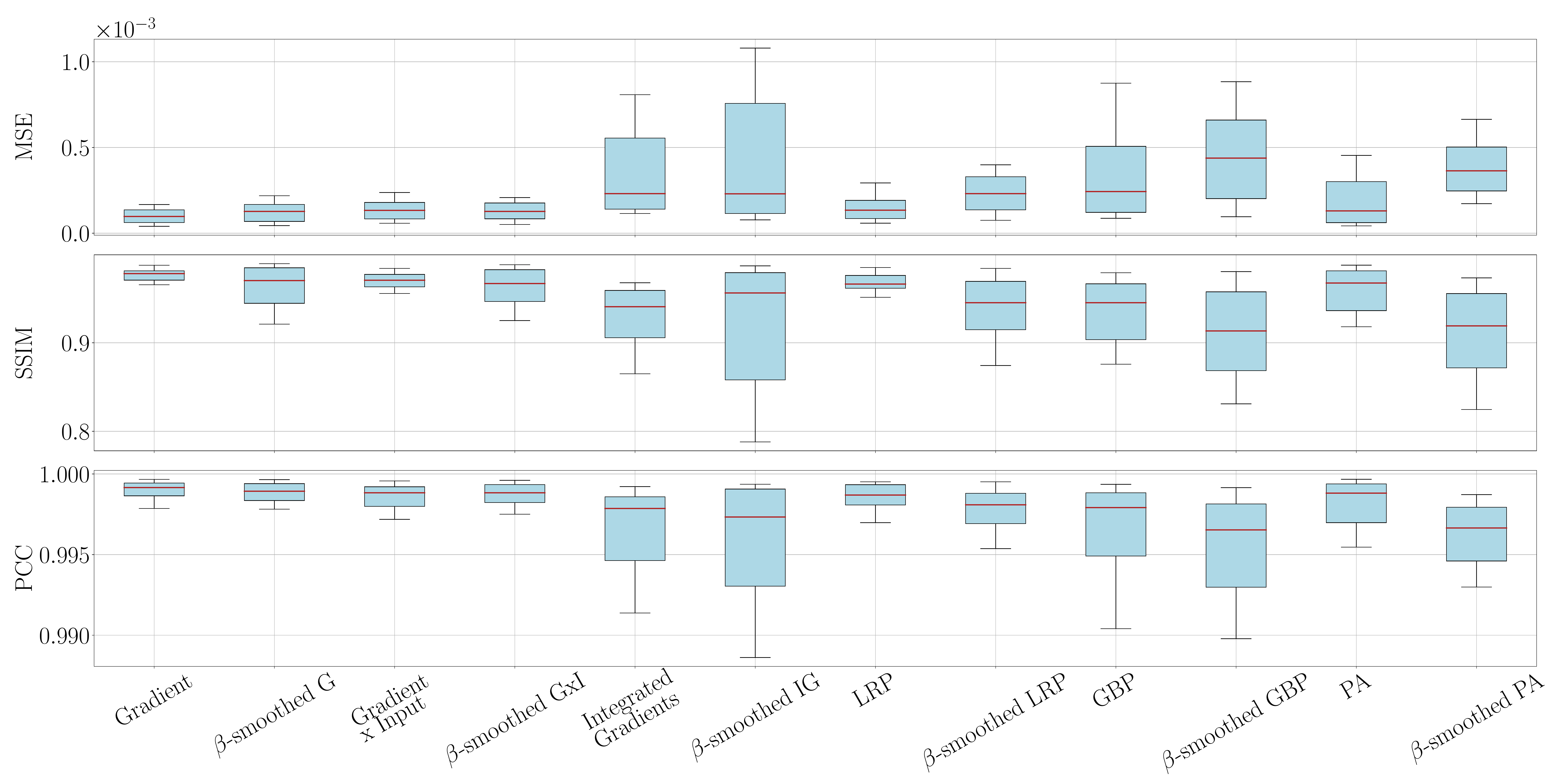}
  \caption{Comparison of Similarities between original and manipulated images. The similarity measures for images for the $\beta$-smoothed explanation methods are of comparable size or slightly worse (higher MSE, lower SSIM and lower PCC) than for the original explanation method, i.e. the manipulations are more visible for the $\beta$-smoothed explanation methods.\label{fig:vanilla_vs_soft_image}}
\end{figure}

In Figure~\ref{fig:vanilla_vs_soft_hm} and Figure~\ref{fig:vanilla_vs_soft_image}, we directly compare the original explanation methods with the $\beta$-smoothed explanation methods. An increase in robustness can be seen for all methods: explanation maps for $\beta$-smoothed explanations have higher MSE and lower SSIM and PCC than explanation maps for the original methods. The similarity measures for the manipulated images are of comparable magnitude.

\begin{figure}[h]
  \centering
  \includegraphics[width=.4\linewidth]{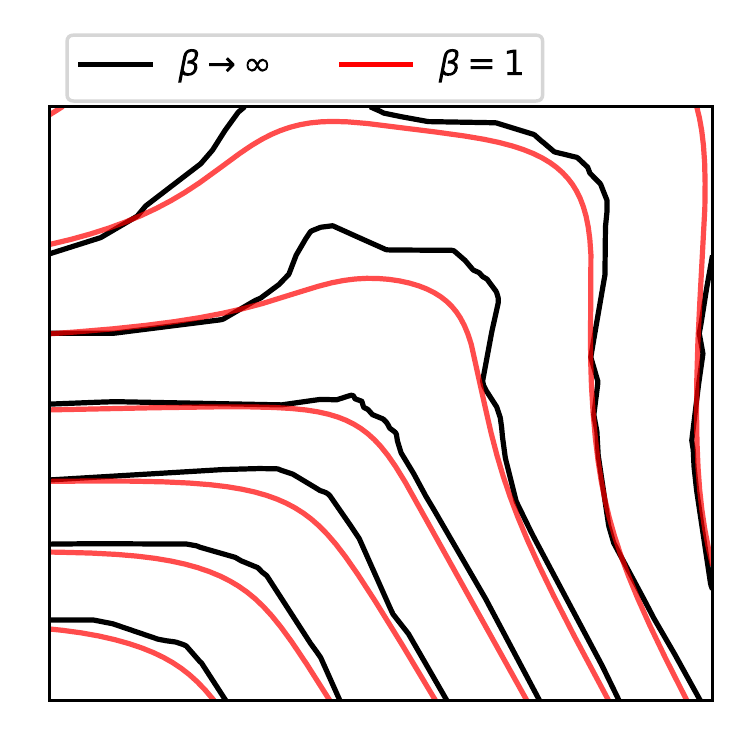}
  \caption{%
    Contour plot of a 2-Layer Neural Network %
    $f(x) = V^\top \softplus (W^\top x)$ %
    with $x \in [-1,1]^2$, $W \in \R^{2 \times 50}$, $V \in \R^{50}$ %
    and $V_i, W_{ij} \sim \mathbb{U}(-1, 1)$. %
    Using a softplus activation with $\beta = 1$ %
    visibly reduces curvature compared to a ReLU activation with %
    $\beta \rightarrow \infty$.%
    \label{fig:toy_equipots}}
\end{figure}

\FloatBarrier
\section{Proofs}
In this section, we collect the proofs of the theorems stated in the main text.
\subsection{Theorem 1}
\begin{theorem}
Let $f: \mathbb{R}^d \to \mathbb{R}$ be a network with $\softplus_\beta$ non-linearities and $\mathcal{U}_\epsilon(p) = \{ x \in \mathbb{R}^d ; \left\| x - p \right\| < \epsilon \}$ an environment of a point $p\in S$ such that $\mathcal{U}_\epsilon(p) \cap S$ is fully connected. Let $f$ have bounded derivatives $\left\|\nabla f(x)\right\|\ge c$ for all $x\in \mathcal{U}_\epsilon(p) \cap S$. It then follows for all $p_0 \in \mathcal{U}_\epsilon(p) \cap S$ that
\begin{align}
    \left\| h(p) - h(p_0) \right\| \le |\lambda_{max}| \; d_g(p, p_0) \le \beta \, C \, d_g(p, p_0),
\end{align}
where $\lambda_{max}$ is the principle curvatures with the largest absolute value for any point in $\mathcal{U}_\epsilon(p) \cap S$ and the constant $C>0$ depends on the weights of the neural network.
\end{theorem}
\textbf{Proof:} This proof will proceed in four steps. We will first bound the Frobenius norm of the Hessian of the network $f$. From this, we will deduce an upper bound on the Frobenius norm of the second fundamental form. This in turn will allow us to bound the largest principle curvature $|\lambda_{max}| = \max \{ |\lambda_1| \dots |\lambda_{d-1}|\}$. Finally, we will bound the maximal and minimal change in explanation.

\textbf{Step 1:} Let $\softplus^{(l)}(x) = \softplus (W^{(l)} x)$ where $W^{(l)}$ are the weights of layer $l$.\footnote{We do not make the dependence of softplus on its $\beta$ parameter explicit to ease notation.} We note that 
\begin{align}
    \partial_k \softplus(\sum_j W_{ij} x_j) = W_{ik} \, \sigma(\sum_j W_{ij} x_j) \\
    \partial_l \sigma(\sum_j W_{ij} x_j) = \beta \, W_{il} \, g(\sum_j W_{ij} x_j))
\end{align}
where 
\begin{align}
    \sigma(x)=\frac{1}{(1+e^{-\beta x})} \,,  && g(x) = \frac{1}{(e^{\beta x/2}+e^{-\beta x/2})^2} \,.
\end{align}
The activation at layer $L$ is then given by
\begin{align}
    a^{(L)}(x) = (\softplus^{(L)} \circ \dots \circ \softplus^{(1)})(x)
\end{align}
Its derivative is given by
\begin{align*}
    \partial_k a^{(L)}_i &= \sum_{s_2 \dots s_L} W^{(L)}_{i s_L} \sigma\left(\sum_j W^{(L)}_{ij} a^{(L-1)}_j\right) W^{(L-1)}_{s_{L} s_{L-1}} \sigma\left(\sum_j W^{(L-1)}_{s_{L}j} a^{(L-2)}_j\right) \\
    &\dots W^{(1)}_{s_{2}k} \sigma\left(\sum_j W^{(1)}_{s_2 j} x_j\right)
\end{align*}
We therefore obtain
\begin{align}
    \left \| \nabla a^{(L)} \right\| \le \prod_{l=1}^L \left\| W^{(l)} \right\|_F
\end{align}
Deriving the expression for $\partial_k a^{(L)}_i$ again, we obtain
\begin{align*}
    \partial_l \partial_k a^{(L)}_i = &\sum_m \sum_{s_2 \dots s_L} \{ \\
    &W^{(L)}_{i s_L} \sigma\left(\sum_j W^{(L)}_{ij} a^{(L-1)}_j\right) W^{(L-1)}_{s_{L} s_{L-1}} \sigma\left(\sum_j W^{(L-1)}_{ij} a^{(L-2)}_j\right) \\
    & \dots \beta \sum_{\hat{s}_m} W^{(m)}_{s_{m+1} \hat{s}_m} W^{(m)}_{s_{m+1}s_{m}} g\left(\sum_j W^{(m)}_{s_{m+1} j} a^{(m-1)}_j(x)\right) \partial_{l} a^{(m-1)}_{\hat{s}_m}(x) \\
    & \dots W^{(1)}_{s_2 k} \sigma\left(\sum_j W^{(1)}_{s_2 j} x_j\right) \}
\end{align*}
We now restrict to the case for which the index $i$ only takes a single value and use $|\sigma(\bullet)| \le 1$. The Hessian $H_{ij}=\partial_i \partial_j a^L(x)$ is then bounded by
\begin{align}
    \left \| H \right\|_F  \le \beta \tilde{C}
\end{align}
where the constant is given by
\begin{align}
    \tilde{C} = \sum_m \left \| W^{(L)} \right\|_F \left \| W^{(L-1)} \right\|_F \dots \left \| W^{(m)} \right\|_F^2 \dots \left \| W^{(1)} \right\|_F^2 \,.
\end{align}

\textbf{Step 2:} Let $e_1 \dots e_{d-1}$ be a basis of the tangent space $T_p S$. Then the second fundamental form for the hypersurface $f(x)=c$ at point $p$ is given by
\begin{align}
    \mathcal{L}(e_i, e_j) &= - \langle D_{e_i} n(p) , e_j \rangle \\
    &= - \langle D_{e_i} \frac{\nabla f(p)}{\left\|\nabla f(p)\right\|} , e_j \rangle
    &= - \frac{1}{\left\|\nabla f(p)\right\|} \langle H[f] e_i ,e_j \rangle + (\dots) \langle \nabla f(p), e_j \rangle 
\end{align}
We now use the fact that $\langle \nabla f(p), e_j \rangle = 0$, i.e. the gradient of $f$ is normal to the tangent space. This property was explained in the main text. This allows us to deduce that
\begin{align}
    \mathcal{L}(e_i, e_j) = - \frac{1}{\left\|\nabla f(p)\right\|} H[f]_{ij} \,.
\end{align}
\textbf{Step 3:} The Frobenius norm of the second fundamental form (considered as a matrix in the sense of step 2) can be written as
\begin{align}
    \left\| \mathcal{L} \right\|_F = \sqrt{\lambda_1^2 + \dots + \lambda_{d-1}^2} \,,
\end{align}
where $\lambda_i$ are the principle curvatures. This property follows from the fact that the second fundamental form is symmetric and can therefore be diagonalized with real eigenvectors, e.g. the principle curvatures. Using the fact that the derivative of the network is bounded from below, $\left \| \nabla f(p) \right\| \ge c$, we obtain
\begin{align}
|\lambda_{max}| \le \beta \; \frac{\tilde{C}}{c}  \,. \label{eq:boundcurv}
\end{align}
\textbf{Step 4:} For $p, p_0 \in \mathcal{U}_\epsilon(p) \cap S$, we choose a curve $\gamma$ with $\gamma(t_0)=p_0$ and $\gamma(t)=p$. Furthermore, we use the notation $u(t) = \dot{\gamma}(t)$. It then follows that
\begin{align}
    n(p) - n(p_0) = \int^t_{t_0} \, \frac{d}{dt} \left( n(\gamma(t)) \right) \, dt = \int^t_{t_0} \, D_{u(t)} n(\gamma(t)) \, dt
\end{align}
Using the fact that $ D_{u(t)} n(\gamma(t)) \in T_{\gamma(t)}S$ and choosing an orthonormal basis $e_i(t)$ for the tangent spaces, we obtain
\begin{align}
    \int^t_{t_0} \, D_{u(t)} n(\gamma(t)) \, dt &= \int^t_{t_0} \, \sum_j  \langle e_j(t), D_{u(t)} n(\gamma(t)) \rangle \, e_j(t) \, dt \\
    &= \int^t_{t_0} \, \sum_j  \mathcal{L}( e_j(t), u(t) )  \, e_j(t) \, dt
\end{align}
The second fundamental form $\mathcal{L}$ is bilinear and therefore
\begin{align}
   \int^t_{t_0} \, \sum_{i}  \mathcal{L}( e_j(t), u(t) )  \, e_j(t) \, dt = \int^t_{t_0} \, \sum_{i, j}  \mathcal{L}( e_j(t), e_i(t) ) \,u^i(t)  \, e_j(t) \, dt
\end{align}
We now use the notation $\mathcal{L}_{ij}(t) = \mathcal{L}( e_j(t), e_i(t) )$ and choose its eigenbasis for $e_i(t)$. We then obtain for the difference in the unit normals:
\begin{align}
    n(p) - n(p_0) =  \int^t_{t_0} \, \sum_{i}  \lambda_i(t)  \,u^i(t)  \, e_i(t) \, dt \,, \label{eq:interm}
\end{align}
where $\lambda_i(t)$ denote the principle curvatures at $\gamma(t)$.
By orthonormality of the eigenbasis, it can be easily checked that
\begin{align*}
    &\langle \sum_{i}  \lambda_i(t)  \,u^i(t)  \, e_i(t), \sum_{j}  \lambda_j(t)  \,u^j(t)  \, e_j(t) \rangle \le  |\lambda_{max}|^2 \, \sum_i u^i(t)^2 \\ &\Rightarrow \left\|\sum_{i}  \lambda_i(t)  \,u^i(t)  \, e_i(t)  \right\| \le |\lambda_{max}| \left\| u(t) \right\|\, 
\end{align*}
Using this relation and the triangle inequality, we then obtain by taking the norm on both sides of \eqref{eq:interm}: 
\begin{align} 
    \left\| n(p) - n(p_0) \right\| \le |\lambda_{max}| \, \int^t_{t_0}   \left\| \dot{\gamma}(t)  \right\|\, dt \,.
\end{align}
This inequality holds for any curve connecting $p$ and $p_0$ but the tightest bound follows by choosing $\gamma$ to be the shortest possible path in $\mathcal{U}_\epsilon(p) \cap S$ with length $\int^t_{t_0}   \left\| \dot{\gamma}(t)  \right\| \, dt$, i.e. the geodesic distance $d_g(p,p_0)$ on  $\mathcal{U}_\epsilon(p) \cap S$. The second inequality of the theorem is obtained by the upper bound on the largest principle curvature $\lambda_{max}$ derived above, i.e. \eqref{eq:boundcurv}.

\subsection{Theorem 2}
\begin{theorem}
For one layer neural networks $g(x)=\relu(w^T x)$ and  $g_{\beta}(x)=\softplus_\beta(w^T x)$, it holds that
\begin{align}
   \mathbb{E}_{\epsilon \sim p_{\beta}} \left[ \nabla g(x-\epsilon) \right] = \nabla g_{\frac{\beta}{\left\| w \right\|}}(x) \,, 
\end{align}
where $p_\beta(\epsilon) = \frac{\beta}{(e^{\beta \epsilon/2}+e^{-\beta \epsilon/2})^2}$.
\end{theorem}
\textbf{Proof:} We first show that
\begin{align}
   \softplus_{\beta}(x) = \mathbb{E}_{\epsilon \sim p_{\beta}} \left[ \relu(x) )\right]\,, \label{eq:scalarsp}
\end{align}
for a scalar input $x$. This follows by defining $p(\epsilon)$ implicitly as 
\begin{align}
    \softplus_\beta( x ) = \int_{-\infty}^{+\infty} p(\epsilon) \, \relu(x-\epsilon) \, \text{d}\epsilon \,.
\end{align}
Differentiating both sides of this equation with respect to $x$ results in 
\begin{align}
    \sigma_\beta(x) = \int_{-\infty}^{+\infty} p(\epsilon) \, \Theta(x-\epsilon) \, \text{d}\epsilon = \int_{-\infty}^{x} p(\epsilon)  \, \text{d}\epsilon \,,
\end{align}
where $\Theta(x)= \mathbb{I}(x>1)$ is the Heaviside step function and $\sigma_\beta(x)=\frac{1}{(1+e^{-\beta x})}$. Differentiating both sides with respect to $x$ again results in
\begin{align}
     p_{\beta}(x) = p(x) \,.
\end{align}
Therefore, \eqref{eq:scalarsp} holds. For a vector input $\vec{x}$, we define the distribution of its perturbation $\vec{\epsilon}$ by
\begin{align}
    p_{\beta}(\vec{\epsilon}) = \prod_i p_{\beta}(\epsilon_i) \,,
\end{align}
where $\epsilon_i$ denotes the components of $\vec{\epsilon}$.
We will suppress any arrows denoting vector-valued variables in the following in order to ease notation. We choose an orthogonal basis such that
\begin{align}
    \epsilon = \epsilon_p \hat{w} + \sum_{i} \epsilon^{(i)}_o \hat{w}^{(i)}_o && \text{with} && \hat{w} \cdot \hat{w}_o^{(i)} = 0 && \text{and} && w =   \left\| w \right\| \hat{w} \,.
\end{align}
This allows us to rewrite 
\begin{align*}
    \mathbb{E}_{\epsilon \sim p_{\beta}} \left[ \relu(w^T (x-\epsilon) )\right] &= \mathbb{E}_{\epsilon \sim p_{\beta}} \left[ \relu(w^T x- \left\| w \right\| \epsilon_p) )\right] \\
    &= \int p_{\beta}(\epsilon_p) \left( \relu(w^T x - \left\| w \right\| \epsilon_p )\right) \text{d}\epsilon_p
\end{align*}
By changing the integration variable to $\epsilon = \left\| w \right\| \epsilon_p$ and using \eqref{eq:scalarsp}, we obtain
\begin{align}
   \softplus_{\frac{\beta}{\left\| w \right\|}}(w^T x) = \mathbb{E}_{\epsilon \sim p_{\beta}} \left[ \relu(w^T (x-\epsilon) )\right]\,,
\end{align}
The theorem then follows by deriving both sides of the equation with respect to $x$.

\FloatBarrier
\section{Additional examples for VGG}
\begin{figure}[h]
  \centering
  \includegraphics[width=1.0\linewidth]{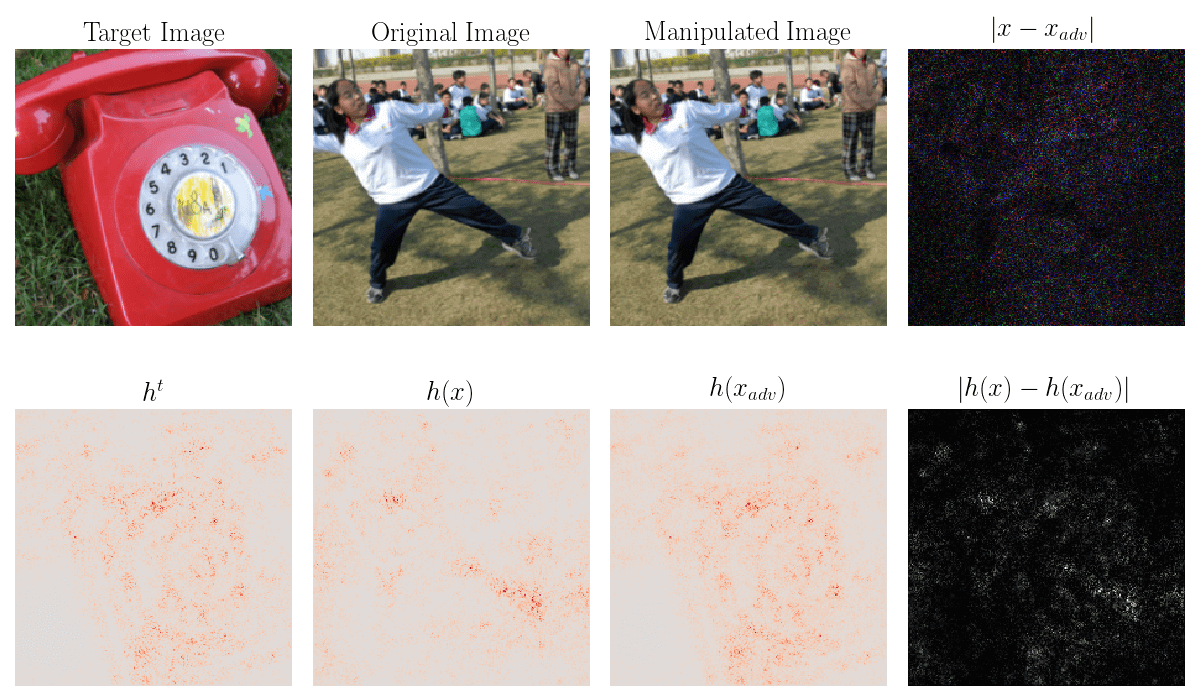}\vspace{1cm}
     \includegraphics[width=1.0\linewidth]{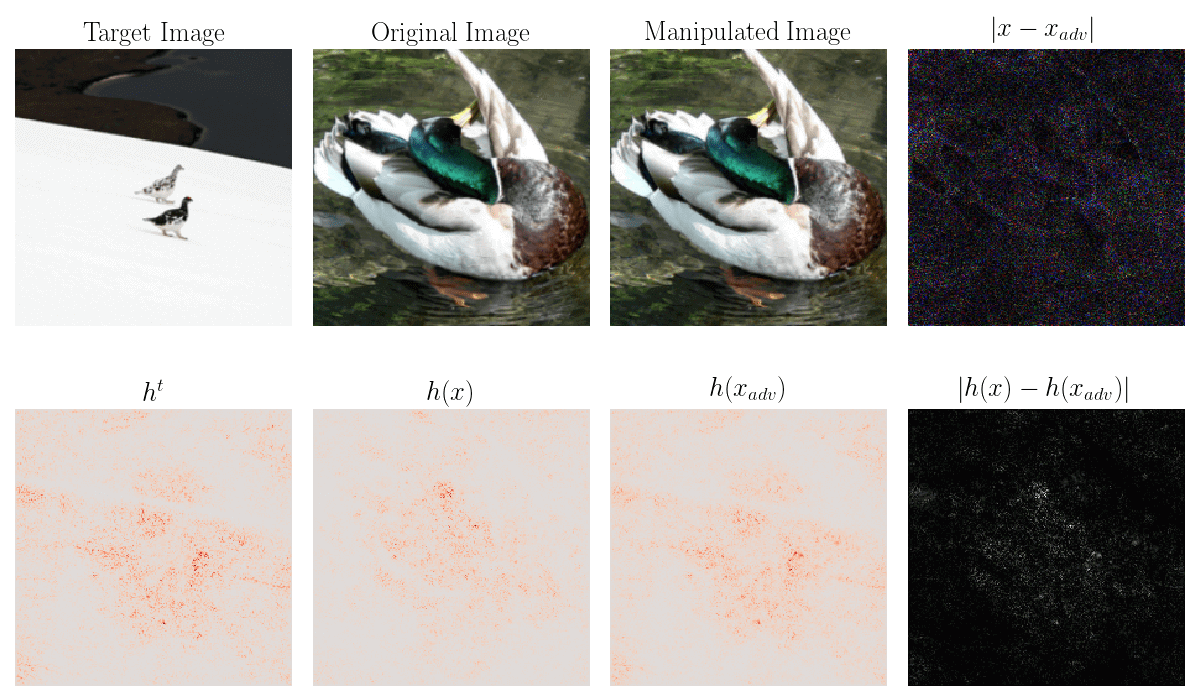}
  \caption{Explanation map produced with the Gradient Explanation Method on VGG. \label{fig:overview_gradient}}
\end{figure}

\begin{figure}[h]
  \centering
  \includegraphics[width=1.0\linewidth]{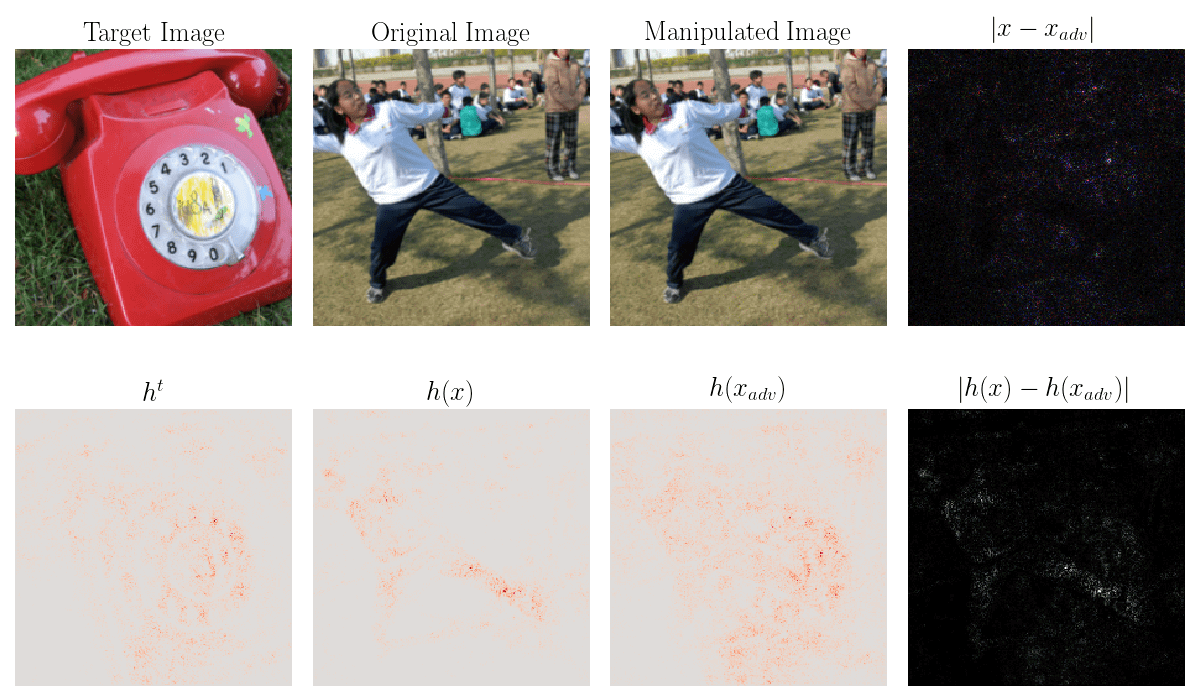}\vspace{1cm}
    \includegraphics[width=1.0\linewidth]{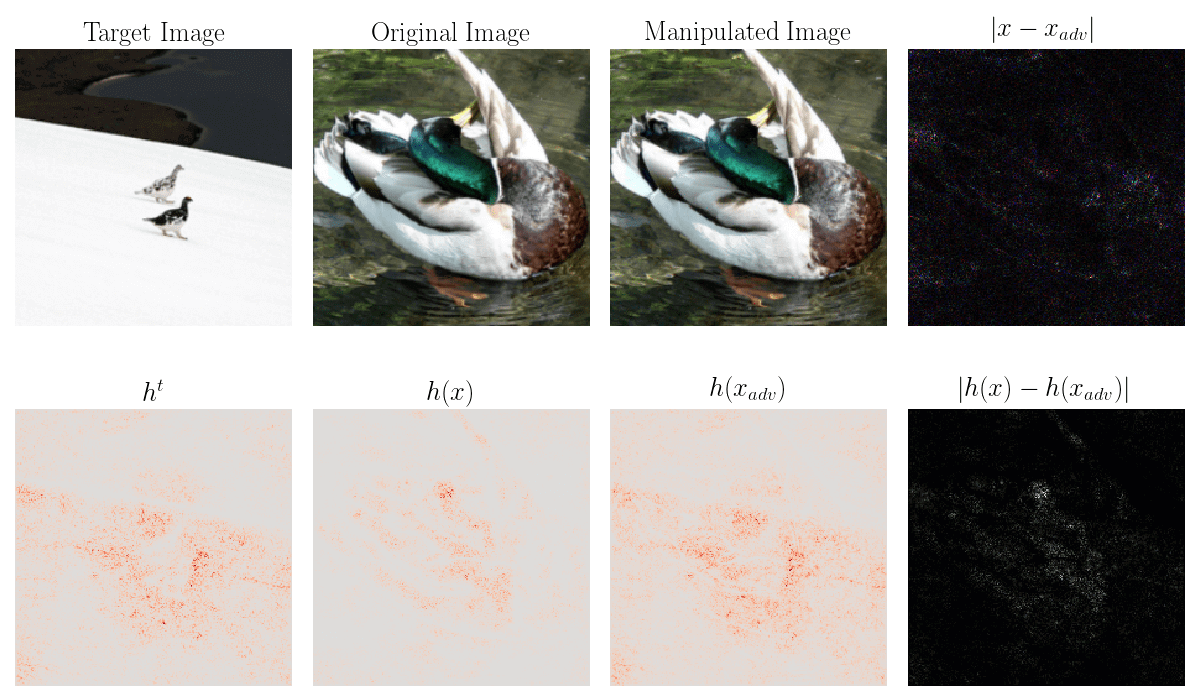}
  \caption{Explanation map produced with the Gradient $\times$ Input Explanation Method on VGG. \label{fig:overview_grad_times_input}}
\end{figure}

\begin{figure}[h]
  \centering
  \includegraphics[width=1.0\linewidth]{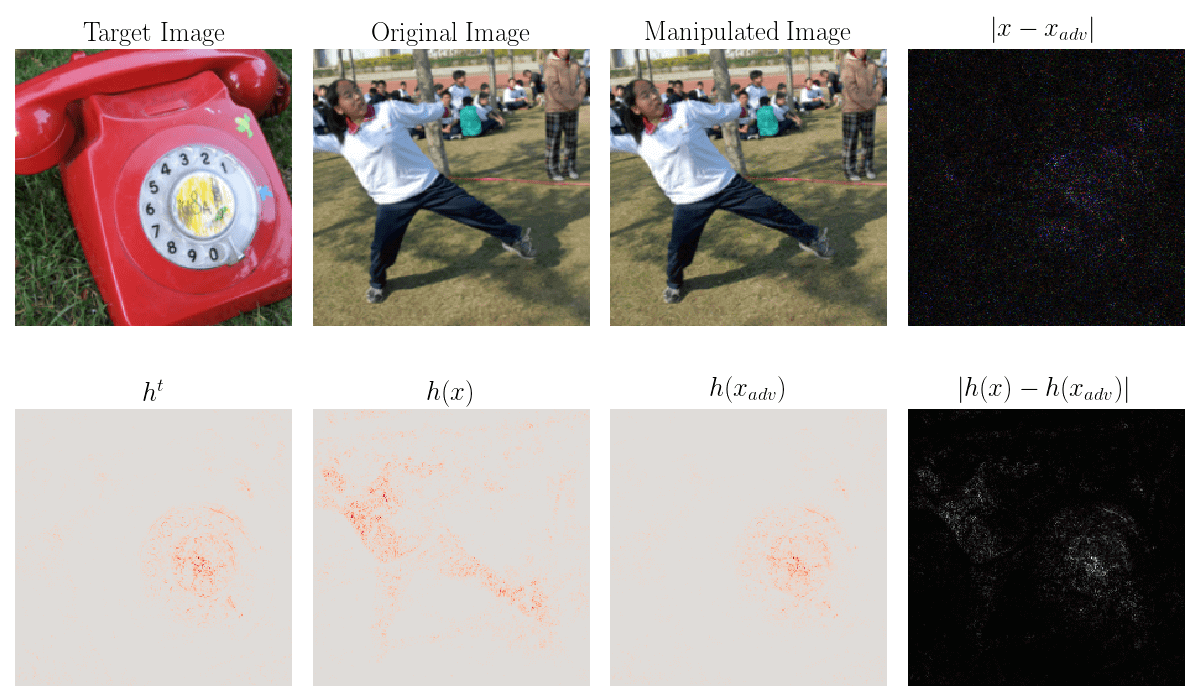}\vspace{1cm}
    \includegraphics[width=1.0\linewidth]{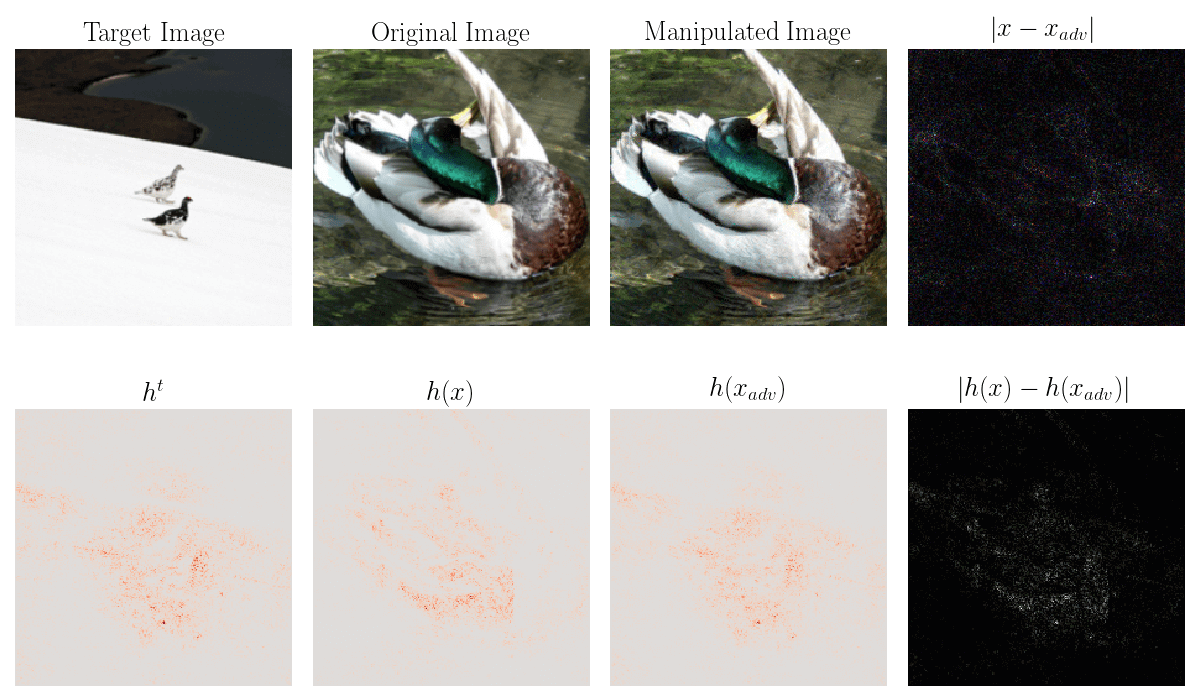}
  \caption{Explanation map produced with the Integrated Gradients Explanation Method on VGG. \label{fig:overview_integrated_grad}}
\end{figure}

\begin{figure}[h]
  \centering
  \includegraphics[width=1.0\linewidth]{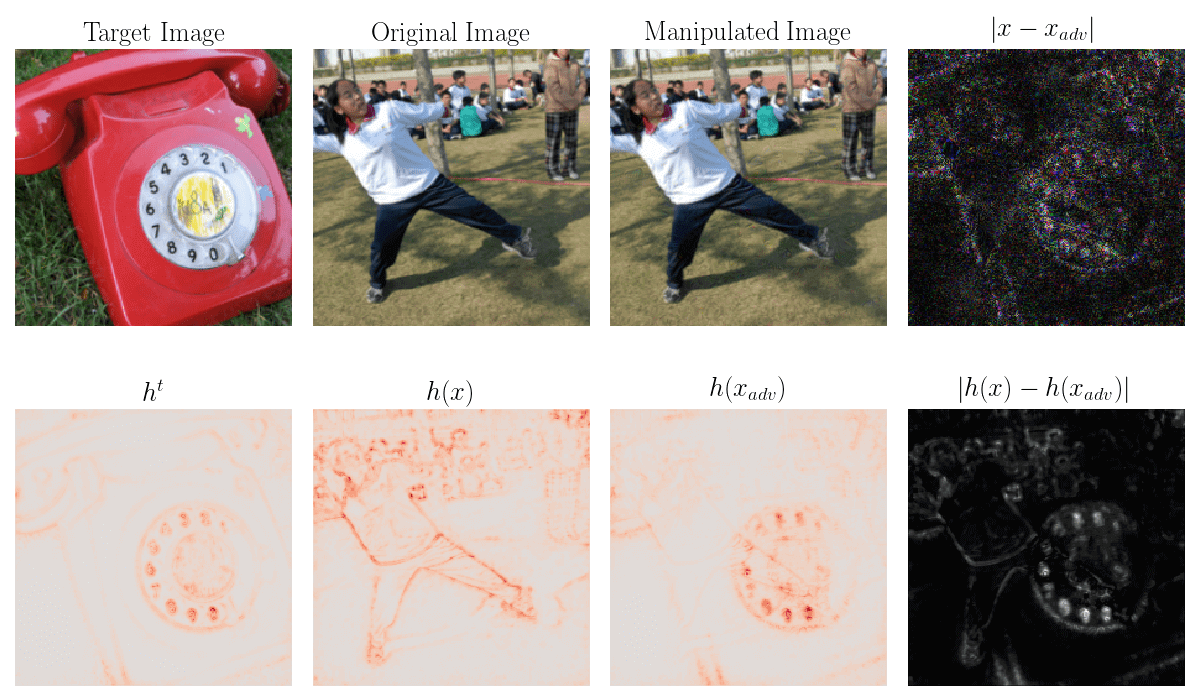}\vspace{1cm}
    \includegraphics[width=1.0\linewidth]{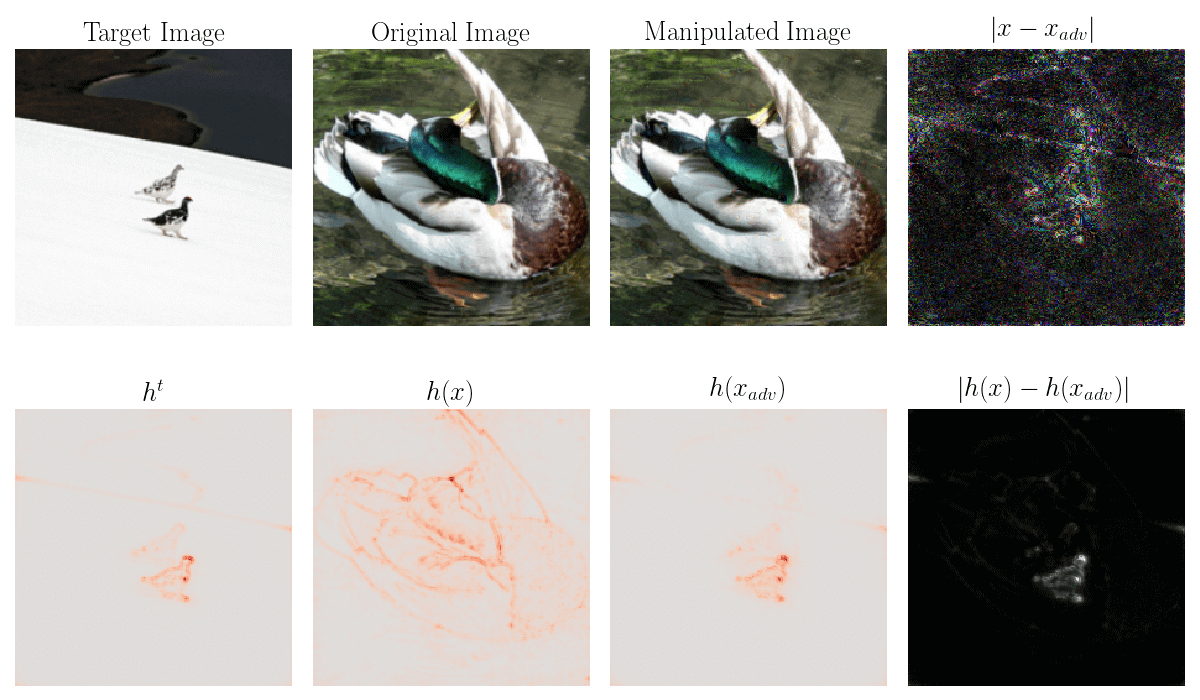}
  \caption{Explanation map produced with the LRP Explanation Method on VGG. \label{fig:overview_lrp}}
\end{figure}

\begin{figure}[h]
  \centering
  \includegraphics[width=1.0\linewidth]{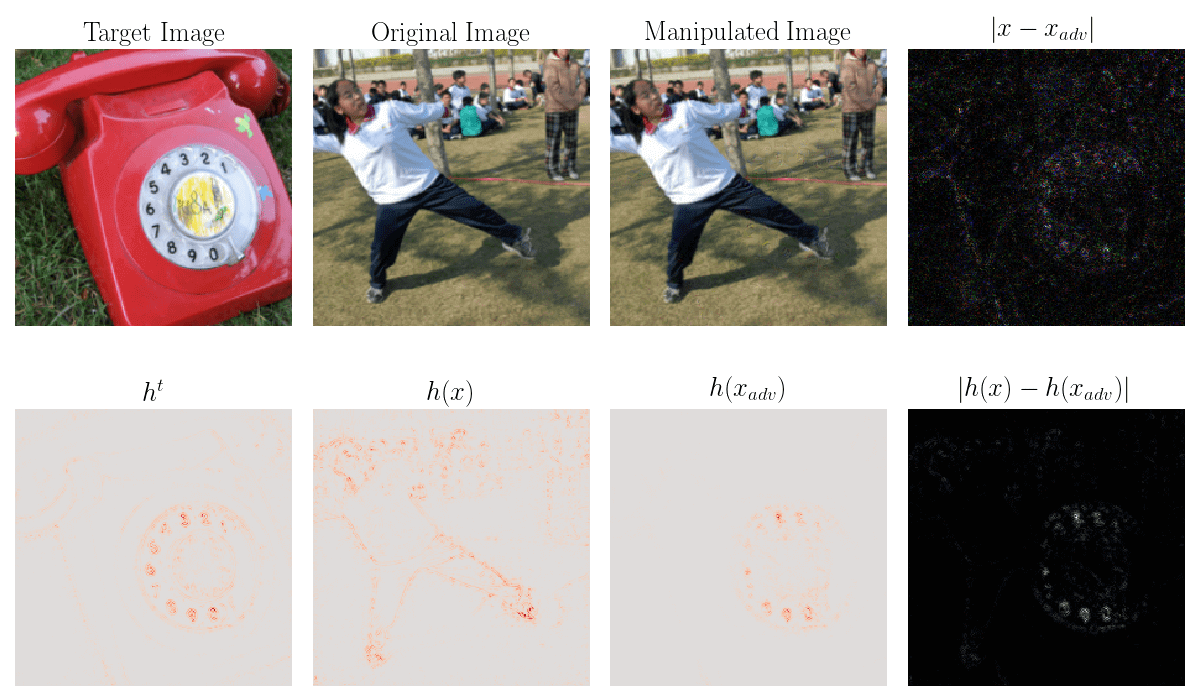}\vspace{1cm}
    \includegraphics[width=1.0\linewidth]{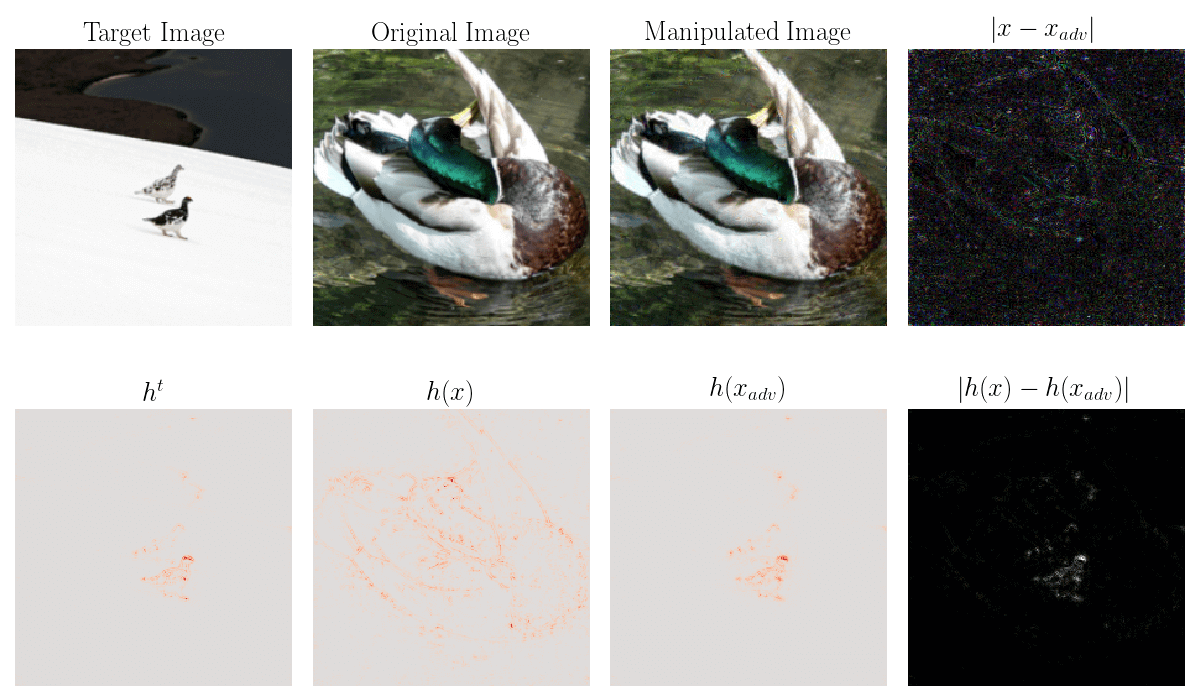}
  \caption{Explanation map produced with the Guided Backpropagation Explanation Method on VGG. \label{fig:overview_guided_backprop}}
\end{figure}

\begin{figure}[h]
  \centering
  \includegraphics[width=1.0\linewidth]{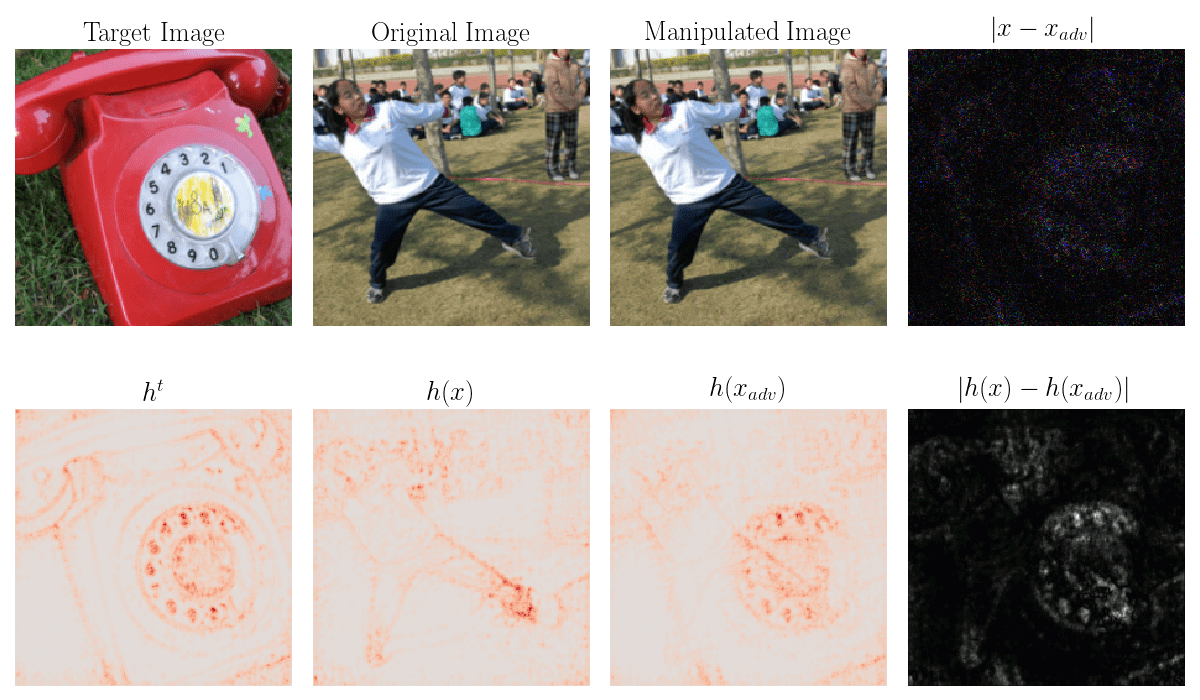}\vspace{1cm}
    \includegraphics[width=1.0\linewidth]{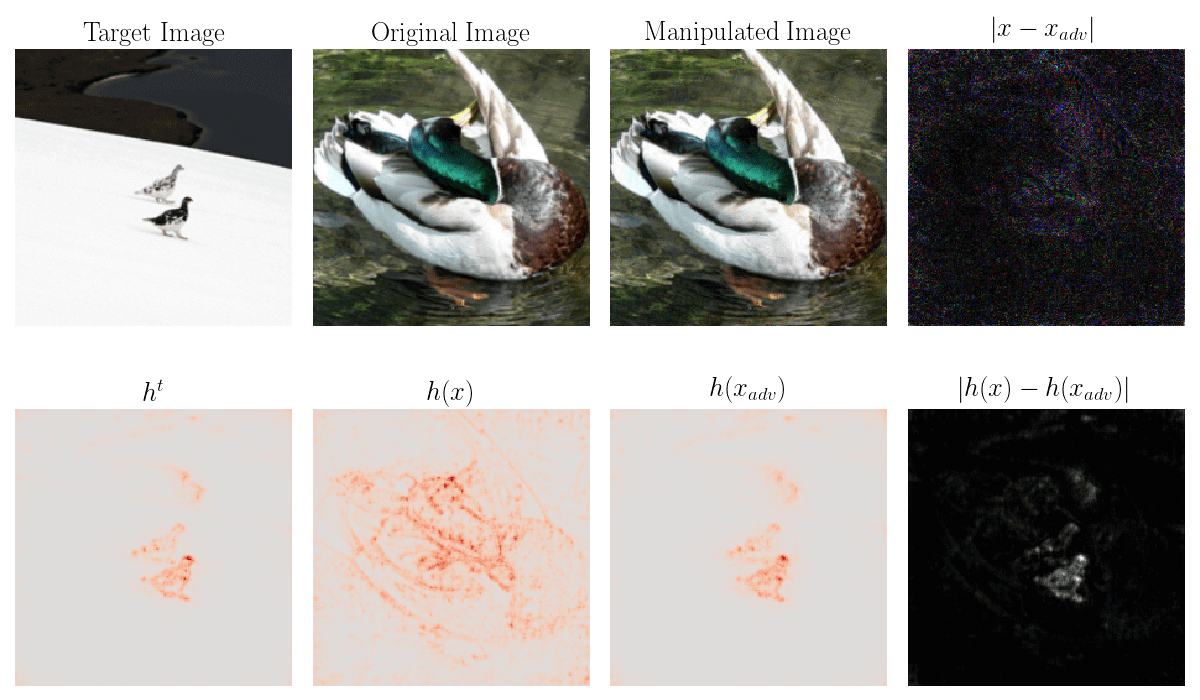}
  \caption{Explanation map produced with the Pattern Attribution Explanation Method on VGG. \label{fig:overview_pattern_attribution}}
\end{figure}

\end{document}